\documentclass{article}

    \PassOptionsToPackage{numbers, sort&compress}{natbib}

\usepackage[preprint]{neurips_2021}

\usepackage[utf8]{inputenc} %
\usepackage[T1]{fontenc}    %
\usepackage{hyperref}       %
\usepackage{url}            %
\usepackage{booktabs}       %
\usepackage{amsfonts}       %
\usepackage{nicefrac}       %
\usepackage{microtype}      %
\usepackage{xcolor}         %
\usepackage{amsmath}

\usepackage{graphicx}
\usepackage{subfigure}
\usepackage{wrapfig}
\usepackage{xspace}

\usepackage[ruled,vlined]{algorithm2e}

\newcommand{\codistillation}{codistillation\xspace}

\newcommand{\allreduce}{\texttt{all\_reduce}\xspace}

\title{A Closer Look at Codistillation \\ for Distributed Training}

\author{Shagun~Sodhani\\
Facebook AI Research\\
Montreal, QC, Canada\\
\texttt{sodhani@fb.com}
\And
Olivier~Delalleau\\
Facebook AI Research\\
Montreal, QC, Canada\\
\texttt{odelalleau@fb.com}
\And
Mahmoud~Assran\\
Mila -- Quebec AI Institute\\
ECE Department, McGill University\\
Facebook AI Research\\
Montreal, QC, Canada\\
\texttt{massran@fb.com}
\And
Koustuv~Sinha \\
Mila -- Quebec AI Institute\\
School of Computer Science\\
McGill University\\
Facebook AI Research\\
Montreal, QC, Canada\\
\texttt{koustuvs@fb.com}
\And
Nicolas~Ballas\\
Facebook AI Research\\
Montreal, QC, Canada\\
\texttt{ballasn@fb.com}
\And
Michael~Rabbat\\
Mila -- Quebec AI Institute\\
Facebook AI Research\\
Montreal, QC, Canada\\
\texttt{mikerabbat@fb.com}}

\begin{document}

\maketitle

\begin{abstract}
Data parallel training is the most widely used approach to speed up neural network training by leveraging the compute resources of multiple devices (e.g., multiple GPUs). Increasing the number of devices brings more computational power but also incurs additional communication overhead because typical implementations synchronize the model parameter copies across all devices at every step. Codistillation is a complementary approach that aims to synchronize the functions represented by the neural networks at different devices without necessarily requiring that all device have precisely the same parameters, leveraging the fact that over-parameterized neural nets can generally represent the same function in many different ways. In this work, we demonstrate that scaling training by incorporating $2$-way codistillation (i.e., codistilling two models, using $X$ devices per model) consistently achieves comparable accuracy to standard data parallel training using $2X$ devices, while communicating up to $1000\times$ fewer bits per iteration. We also reveal potential challenges with scaling to $n$-way codistillation ($n > 2$) and show how these challenges relate to problem-specific characteristics of the model architecture and the dataset.
\end{abstract}

\section{Introduction}

Several recent improvements in the performance of machine learning models can be attributed to scaling the training of neural network models \citep{resnet_he2016deep_residual_learning_for_image_recognition, goyal2017imagenet_in_an_hour, transformer_vaswani2017attention_is_all_you_need, devlin2018bert, shoeybi2019megatron, huang2019gpipe, kaplan2020scaling_laws_for_neural_language_models, lepikhin2020gshard, gpt3_brown2020language_models_are_few_shot_learners}. Most approaches for scaling up training leverage some form of data parallelism (using multiple workers to compute gradients on different training samples in parallel), and the most common approach to data-parallel training is synchronous first-order optimization.

In synchronous data-parallel, several devices (e.g., GPUs) are used to accelerate training. Each device holds a copy of the model being trained, and the copies of these models are kept synchronized throughout training. In one step of a typical implementation, every device computes a gradient using different samples, the gradients are averaged across all devices (e.g., via an \allreduce operation), and then each device locally performs an optimizer step using the average gradient. Since the devices use the average gradient in the optimizer step, and assuming the parameters at every device are initialized to the same value, the parameters remain synchronized after every step. 

Increasing the number of devices, while keeping the per-device batch size fixed, increases the effective (total) batch size used for updates. This reduces the gradient's variance and may allow increasing the learning rate. In the ideal case, when doubling the number of devices, the learning rate can also be doubled and the total number of updates needed to reach a desired level of accuracy is halved, resulting in a linear speedup. In practice, this scaling has only been observed to hold until reaching a problem-dependent critical batch size, and more generally there are diminishing returns when increasing the batch size~\cite{goyal2017imagenet_in_an_hour,johnson2020adascale}.

Synchronizing the full model parameters at every iteration incurs a substantial communication overhead, and a variety of approaches have been proposed to reduce this overhead, including quantizing or compressing gradients before synchronizing them~\cite{alistarh2017qsgd}, synchronizing periodically rather than on every update~\cite{stich2018local_sgd_converges_fast_and_communicates_little}, and only synchronizing among subsets of devices~\cite{sgp_pmlr-v97-assran19a}. While each of these strategies reduces the communication overhead per update, they also introduce some additional error which can impact the resulting model quality or training time.

It is well-known that due to symmetries inherent to typical neural network structures, overparameterized neural network models can represent the same function in many different ways (i.e., with many different parameter values)~\cite{goodfellow2016deep}. Codistillation~\cite{anil2018large,zhang2018deep} is another approach to distributed training that aims to train models in parallel so that they represent the same function (mapping inputs to predictions), without requiring that the models necessarily have the same parameters. This is accomplished by adding a distillation loss that penalizes the predictions made by one model when they differ from the other models. Codistillation and synchronous data-parallel training are complementary, in that one could codistill $n$ models while using $X$ devices per model.

Previous work~\cite{anil2018large} introduced codistillation for distributed training and demonstrated it can be effective in specific scenarios. In one set of experiments (training a language model on CommonCrawl), the training dataset is so large that training terminates before a single pass is completed over the data. In another experiment (training ResNet50 on ImageNet), multiple passes are made over the training set, but the batch size is larger than the known critical batch size, and the resulting accuracy achieved is thus also lower than what could be achieved using a smaller batch size. Despite these particularities, both sets of experiments serve to demonstrate that codistillation is promising for distributed training. We also note that the experiments in~\cite{anil2018large} focus on $2$-way codistillation; i.e., codistilling only $n=2$ models, with multiple devices per model.

In this paper, we further investigate codistillation for distributed training. First, we consider the typical training scenarios where multiple passes are made over the training set with a total batch size lower than the critical batch size (training ResNet50 on Imagenet and a transformer-based machine translation model). In this setting, we show that $2$-way codistillation can achieve a linear scaling relationship in that when using $2X$ devices per model one can half the number of training updates without loss in accuracy. This is consistent across image classification and machine translation workloads. Moreover, this is achieved while communicating up to $1000\times$ fewer bits per iteration than standard data-parallel training.

We also consider $n$-way codistillation with $n > 2$, and the results are mixed. While it is possible that codistilling $n > 2$ models leads to higher accuracy for some problems, on other problems $n$-way codistillation is no better than $2$-way codistillation. Building on recent work~\cite{allen-zhu_towards_2020} towards understanding ensembling and distillation, we conduct experiments that help elucidate conditions where one may expect $n$-way codistillation to enhance performance; roughly speaking, $n$-way codistillation helps when it is possible for different models to learn distinct sets of features that may be useful for making predictions on the dataset.

\section{Codistillation: Background and Related Work}

Codistillation is proposed as a mechanism for sharing information between multiple models being trained concurrently~\cite{anil2018large, zhang2018deep}. In the typical student-teacher distillation~\cite{hinton2014distilling_the_knowledge_in_a_neural_network}, there are two phases: first, a teacher model is trained using standard supervised learning, and then a student model is trained to predict the outputs of the teacher model while the teacher's parameters are kept fixed. In contrast, when two or more models \textit{codistill}, there is only one phase, and in addition to minimizing the usual supervised loss on the training data, an additional loss term is used to share knowledge between the models by encouraging each model to make similar predictions to the other(s).

\begin{algorithm}[t]
\SetAlgoLined
\LinesNumbered
\DontPrintSemicolon
\SetKwInOut{Input}{Input}
\Input{Loss function $L(y, \hat{y})$ and \codistillation loss function $D(y, y')$}
\Input{Model architecture $f_\theta(x)$ and initial model parameters $\{\theta_i^1 \colon i = 1,\dots,n\}$}
\Input{Number of iterations $K$, learning rates $\{\eta^k\}_{k=1}^K$, and penalty coefficients $\{\alpha^k\}_{k=1}^K$}
\For{$k = 1, \dots, K$} {
    \For{$i = 1, \dots, n$ {\bf in parallel}} {
        $x,y = $ \texttt{get\_next\_minibatch()} \;
        $\theta_i^{k+1} = \theta_i^k - \eta^k \nabla_{\theta_i} \left( L(y, f_{\theta_i^k}(x)) + \alpha^k \frac{1}{n-1} \sum_{j \neq i} D\left(f_{\theta_i^k}(x), f_{\theta_j^k}(x)\right) \right)$ \label{line:codist_model_update} \;
    }
}
\caption{Codistillation}
\label{alg:codistillation}
\end{algorithm}

Codistillation, as described in~\citep{zhang2018deep}, is shown in Algorithm~\ref{alg:codistillation}. Here, $n \ge 2$ models are trained concurrently. The $i$th model is updated on line~\ref{line:codist_model_update} by taking a gradient step to minimize the combination of a standard supervised loss function $L$ (e.g., cross-entropy or MSE) and a distillation-like loss $D$ which penalizes differences between the predictions made by model $i$ and those made by model $j$, averaged over all other models $j \ne i$. \citet{zhang2018deep} and \citet{anil2018large} both report using Kullback-Liebler (KL) divergence for $D$ in their experiments, and they do not explicitly include a penalty parameter $\alpha^k$, instead (implicitly) taking $\alpha^k = 1$ for all $k$.

\section{Implementation Options and Communication Overhead}
\label{sec:implemnentation_options}
There are multiple ways that Algorithm~\ref{alg:codistillation} may be implemented, and the implementation generally impacts the communication overhead. One approach, suggested in~\cite{anil2018large}, is to exchange model checkpoints. Then at each iteration, a device implementing updates for one model (e.g., $i=1$) will compute $n$ forward passes, one for $f_{\theta_i^k}(x)$ and $n-1$ for $f_{\theta_j^k}(x)$, $j\ne i$, and the device will only compute the backward pass with respect to $\theta_i$ and update those parameters. Communicating model checkpoints is expensive, so~\citet{anil2018large} propose to only exchange checkpoints periodically, e.g., after every 50 updates. Consequently, the predictions from model $j$ used as targets in the distillation loss when updating model $i$ will be based on a stale copy of model $j$, but it is argued that predictions change more slowly than model parameters during training, so codistillation should be reasonably tolerant to staleness~\cite{anil2018large}, and hence is amenable to asynchronous implementation. Moreover, the communication of checkpoints can be overlapped with other operations, and the new model swapped in at the next iteration after it has been received.

An alternative implementation has each device compute a single forward pass and then communicate the predictions. This approach requires that the devices training different models use coordinated sampling so that they process the same mini-batch $(x,y)$. Communicating predictions typically requires many fewer bits than communicating the entire model. However, in addition to requiring coordinated sampling, this approach introduces a synchronization point at every update, since a device cannot compute the backward pass until it has received all predictions needed to compute the distillation loss. For this reason, one may consider only periodically communicating predictions to reduce overhead, and then omit the distillation loss term in line~\ref{line:codist_model_update} on iterations where they are not communicated.

\newcommand{\bmodel}{b_{\text{model}}}
\newcommand{\bpred}{b_{\text{predictions}}}
\newcommand{\Car}{C_{\text{AR}}}
Let $\bmodel$ denote the number of bits required to represent the model parameters, let $\bpred$ denote the number of bits to represent model predictions on a single training sample, and let $B$ denote the per-device batch size. For example, a ResNet50 model predicting 1000 classes and using 32-bit floating point will have $\bmodel = 8 \times 10^8$ bits, and $\bpred = 3.2 \times 10^4$. In an optimized implementation of \allreduce (e.g., ring-based or tree-based), each device communicates $\Car = 2 \bmodel$ bits per iteration. Codistillation with checkpoints communicated every $T$ iterations has every device communicate $(n-1) \bmodel / T$ bits per iteration, on average, which can be smaller than $\Car$ if $(n-1) / T < 2$, e.g., if $n$ is small and $T$ is sufficiently large. Codistillation with predictions communicated every $T$ iterations has every device communicate $(n-1) \bpred B / T$ bits per iteration, on average.

\begin{figure}
    \begin{center}
        \includegraphics[width=0.45\textwidth]{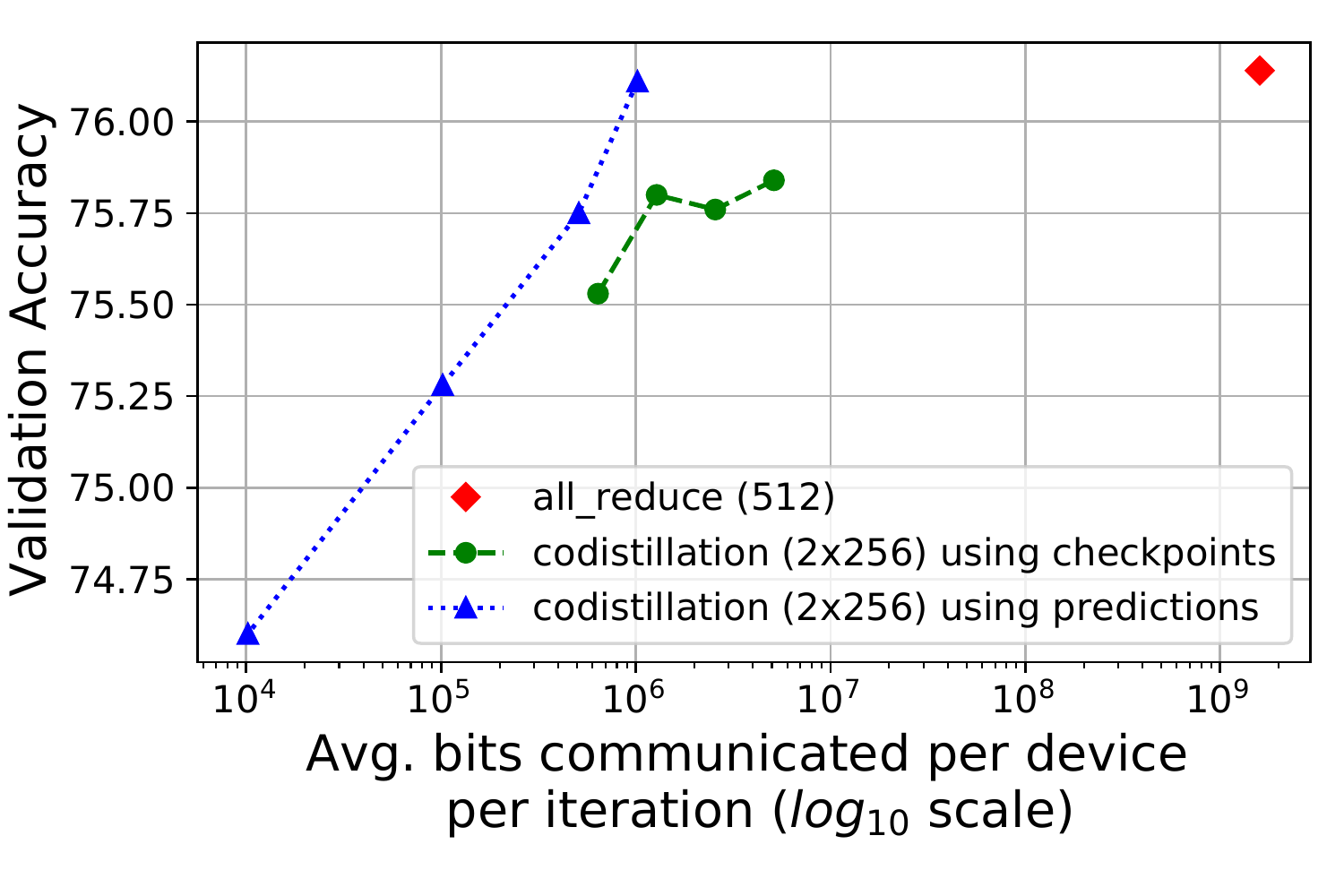}
    \end{center}
    \vspace{-10pt}
\caption{Training ResNet50 on ImageNet using 2-way codistillation with batch size 256 per model. We investigate different ways of implementing codistillation, either by periodically communicating model checkpoints (parameters) or periodically communicating predictions. Different points on the codistillation curves correspond to different communication periods; when communicating predictions, from right to left we consider communicating every 1, 5, 10, and 100 iterations, and when communicating checkpoints we consider every 625, 1250, 2500 and 5000 iterations. Compared to standard data-parallel training (\allreduce), codistillation with predictions achieves comparable top-1 validation accuracy (76.1\%) while communicating $1000\times$ fewer bits per iteration. Following~\cite{goyal2017imagenet_in_an_hour}, we use batch size 32 per GPU, so all methods use 16 GPUs in total, and all methods train for the same number of iterations. For additional information about hyperparameters and experimental setup see Section~\ref{subsec:2way_imagenet}}.
\label{fig:comparing_bits}
\vspace{-10 pt}
\end{figure}

Figure~\ref{fig:comparing_bits} illustrates the accuracy-communication trade-off when training a ResNet50 on ImageNet. All methods train for the same number of updates per model, corresponding to 90 epochs of standard data-parallel training. Following~\cite{goyal2017imagenet_in_an_hour}, we use batch size 32 per GPU, and thus all methods use 16 GPUs in total (8 GPUs per model in 2-way codistillation). In codistillation, each model is trained using 8 GPUs that reside on the same server. Since communication between devices on different servers is much more expensive than between devices on the same server, the figure only counts communication between servers. Additional information about the experimental setup, including other hyperparameters, is discussed in Section~\ref{subsec:2way_imagenet}. For the remainder of the paper, unless otherwise mentioned, codistillation refers to the implementation communicating predictions.

\section{Two-way Codistillation}

The previous section focused on various design tradeoffs arising when implementing codistillation. Next, in this section we demonstrate linear scaling of two-way codistillation in terms of number of iterations to achieve a target accuracy. Initially we focus on training various image classification models on ImageNet, and then we also discuss a machine translation task.

\subsection{Image Classification on ImageNet}
\label{subsec:2way_imagenet}

\citet{goyal2017imagenet_in_an_hour} demonstrate that \allreduce-based data-parallel training can scale linearly when training ResNet50 on ImageNet. When doubling the number of GPUs, the effective (total) batch size per iteration doubles, so to achieve linear scaling the number of total iterations is reduced by half. In order to halve the iterations without loosing accuracy, a number of techniques are introduced in~\cite{goyal2017imagenet_in_an_hour}, with the main ones being learning rate warm-up and linear learning rate scaling (i.e., the peak learning rate increases in proportion to the total batch size).

We begin by adopting a similar training setup. Our \allreduce baseline uses the same hyperparameters, including the step-wise learning rate schedule, suggested in~\cite{goyal2017imagenet_in_an_hour}, and after 90 epochs it achieves a top-1 validation accuracy of 76.13\% (as shown in~\cite{goyal2017imagenet_in_an_hour}) .

\begin{figure}[t]
\centering     %
\subfigure[Training Loss]{\label{fig:compare_ar_with_codistillation_anil_et_al_with_codistillation_proposed_resnet50_train_average_loss}\includegraphics[width=0.33\textwidth]{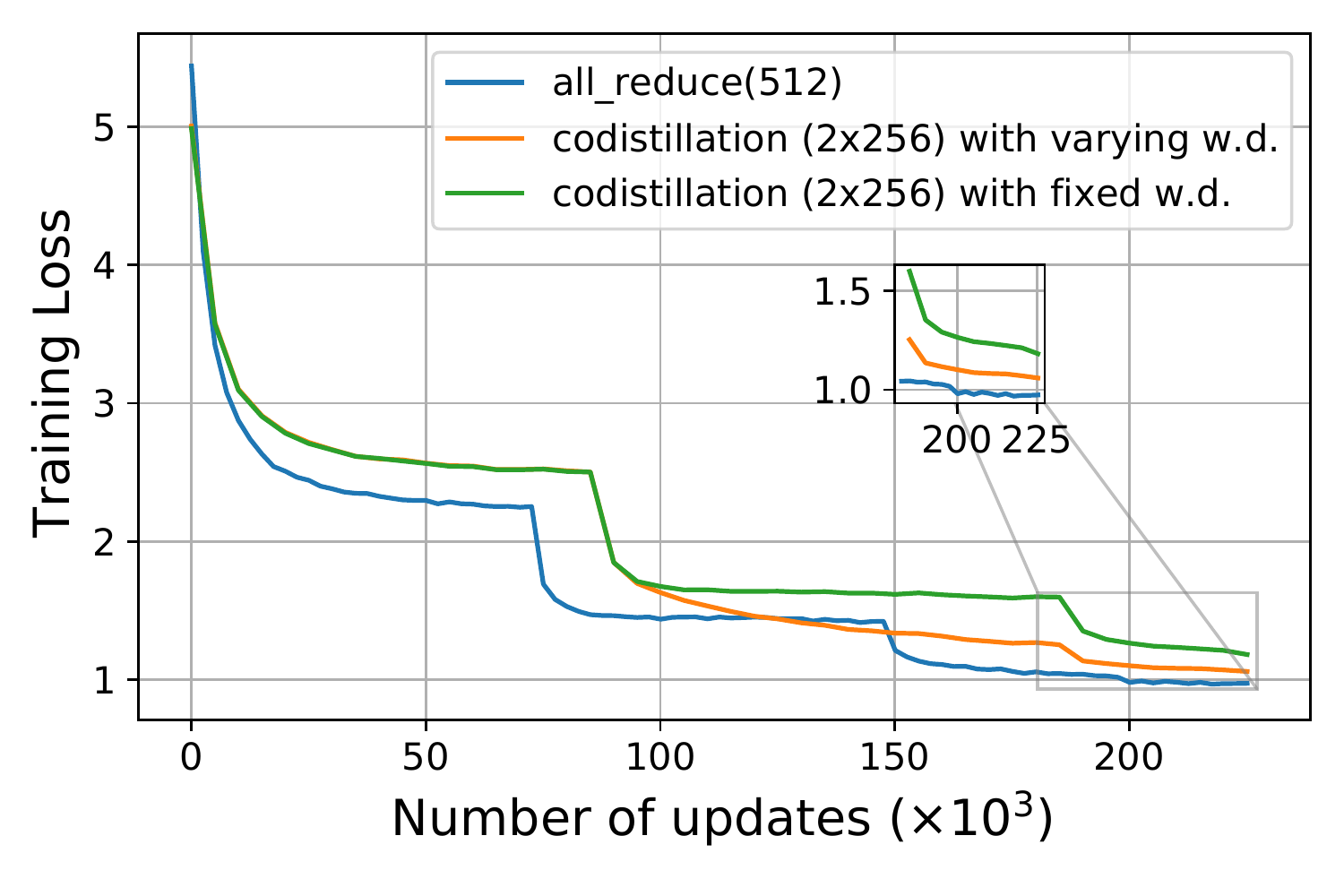}}
\subfigure[Top-1 Validation Accuracy]{\label{fig:compare_ar_with_codistillation_anil_et_al_with_codistillation_proposed_resnet50_validation_top_1_accuracy}\includegraphics[width=0.33\textwidth]{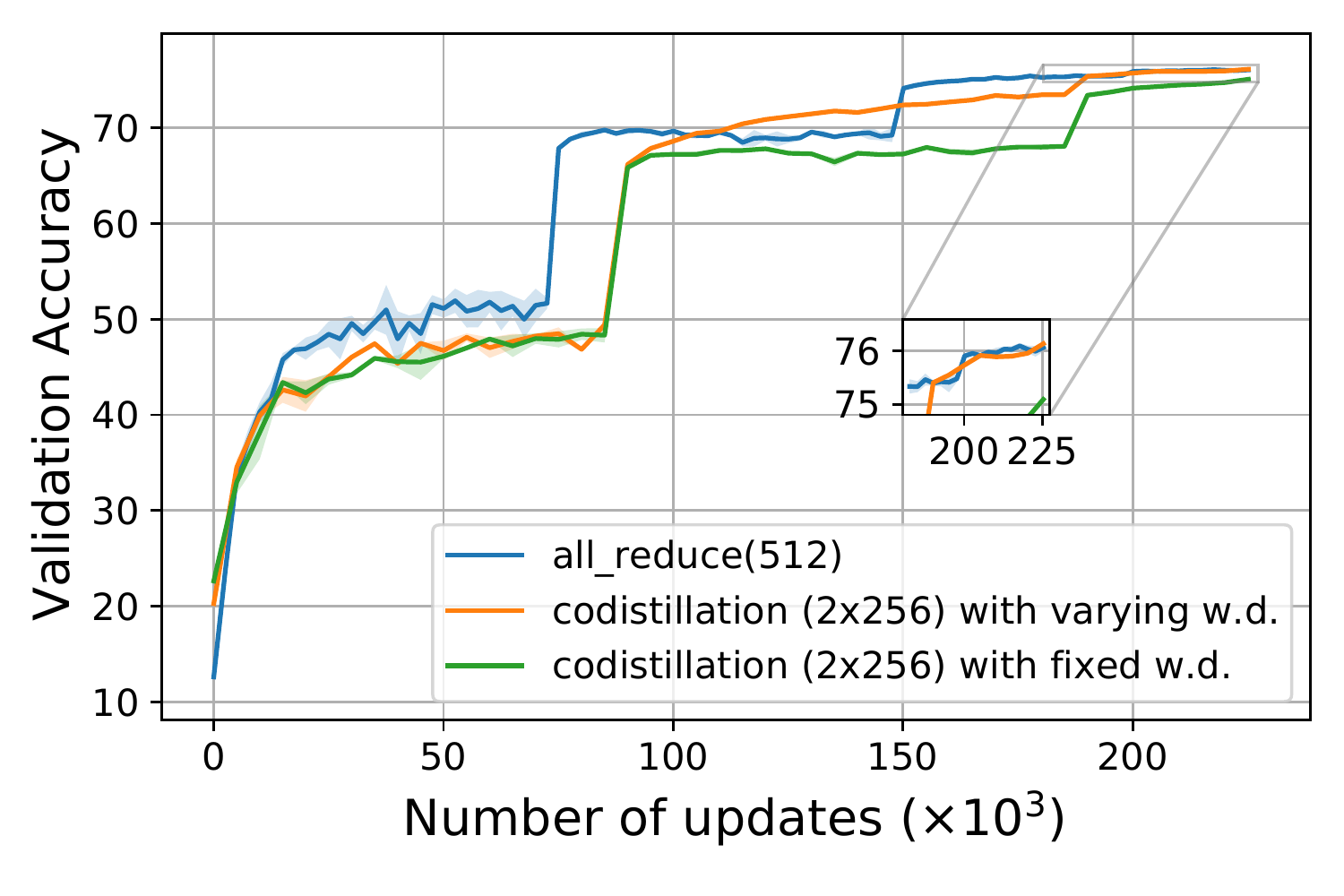}}
\subfigure[Top-1 Validation Accuracy]{\label{fig:compare_codistillation_for_differernt_batch_sizes_resnet50_validation_top_1_accuracy}\includegraphics[width=0.30\textwidth]{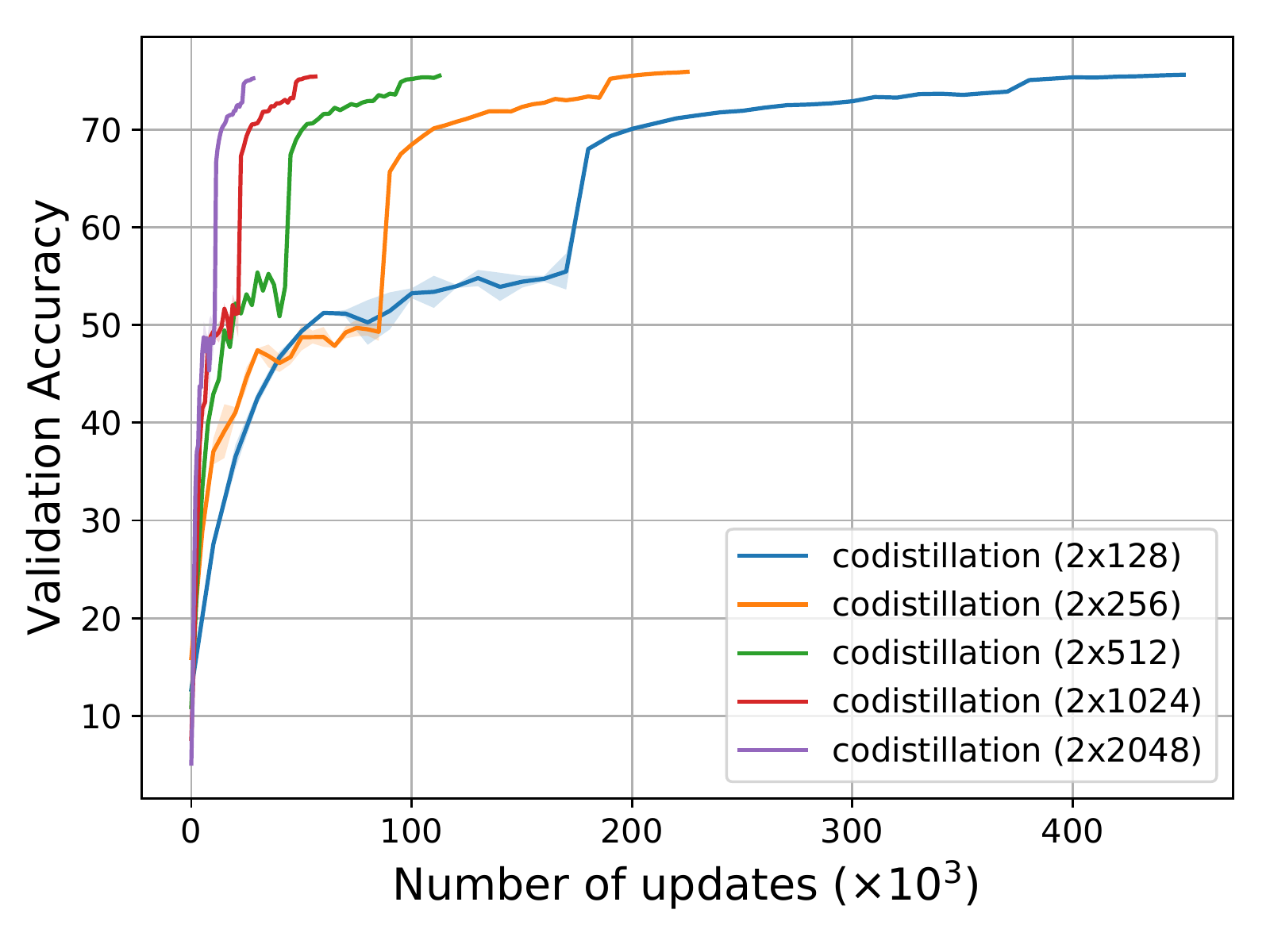}}
\caption{(a) and (b): Comparing the performance of ResNet50 models trained using \allreduce and \codistillation on the ImageNet dataset. We codistill two models using batch size 256 for each, and the model trained with \allreduce uses batch size 512. All experiments are run on 16 GPUs. We report the (a) training loss and (b) top-1 validation accuracy. We observe that compared to \allreduce, the model trained with \codistillation (with fixed weight decay or w.d.) underfits, and obtains a higher training loss and lower top-1 accuracy. Using a variable weight decay (decaying over time) narrows the performance gap, with the \allreduce model only marginally better than the \codistillation model. (c): Codistillation scales well across multiple values of batch size per model. Each time we double the batch size per model, we scale the learning rate schedule by a factor of two and perform half the number of updates. We report the validation accuracy across a wide range of batch sizes.}
\label{fig:compare_ar_with_codistillation_anil_et_al_with_codistillation_proposed_and_scale_experiments}
\end{figure}

When training with codistillation, we adopt the same learning rate warm-up and linear learning rate scaling strategies. Figure~\ref{fig:compare_ar_with_codistillation_anil_et_al_with_codistillation_proposed_and_scale_experiments}(a) and (b) show training and validation curves comparing \allreduce with 2-way codistillation. For 2-way codistillation we also use a step-wise learning rate strategy---decreasing the learning rate three times, by $0.1$ each time---and we optimize the schedule for codistillation, finding that it is better to decrease the learning rate later in training.

In order to achieve performance on par with \allreduce, we also find that it is necessary to impose a schedule on L2 regularization for codistillation. This is because codistillation also acts as a regularizer, and while some L2 regularization helps stabilize the early phase of training, if it remains constant during training then this leads to over-regularization and lower validation accuracy. Thus, we set the L2 regularization parameter to $5 \times 10^{-4}$ initially (as in~\cite{goyal2017imagenet_in_an_hour}), and it is reduced to $10^{-5}$ after the first learning rate decay, and to $0$ after the second learning rate decay.

Figure~\ref{fig:compare_ar_with_codistillation_anil_et_al_with_codistillation_proposed_and_scale_experiments}(c) shows that the possibility to reduce the number of training iterations by half when doing 2-way codistillation is not specific to the specific case of using batch size 256 per model, and in fact the same performance holds across a range of batch sizes per model (corresponding to different numbers of GPUs per model). In all cases, the number of iterations per model is scaled in proportion to the number of GPUs (and hence the batch size). See Appendix~\ref{app_subsec:codistillation_vs_large_batch_training} for additional related results.

\begin{figure}[t]
\centering     %
\subfigure[ResNet50]{\label{fig:compare_ar_with_codistillation_cosine_resnet50_validation_top_1_accuracy}\includegraphics[width=0.48\textwidth]{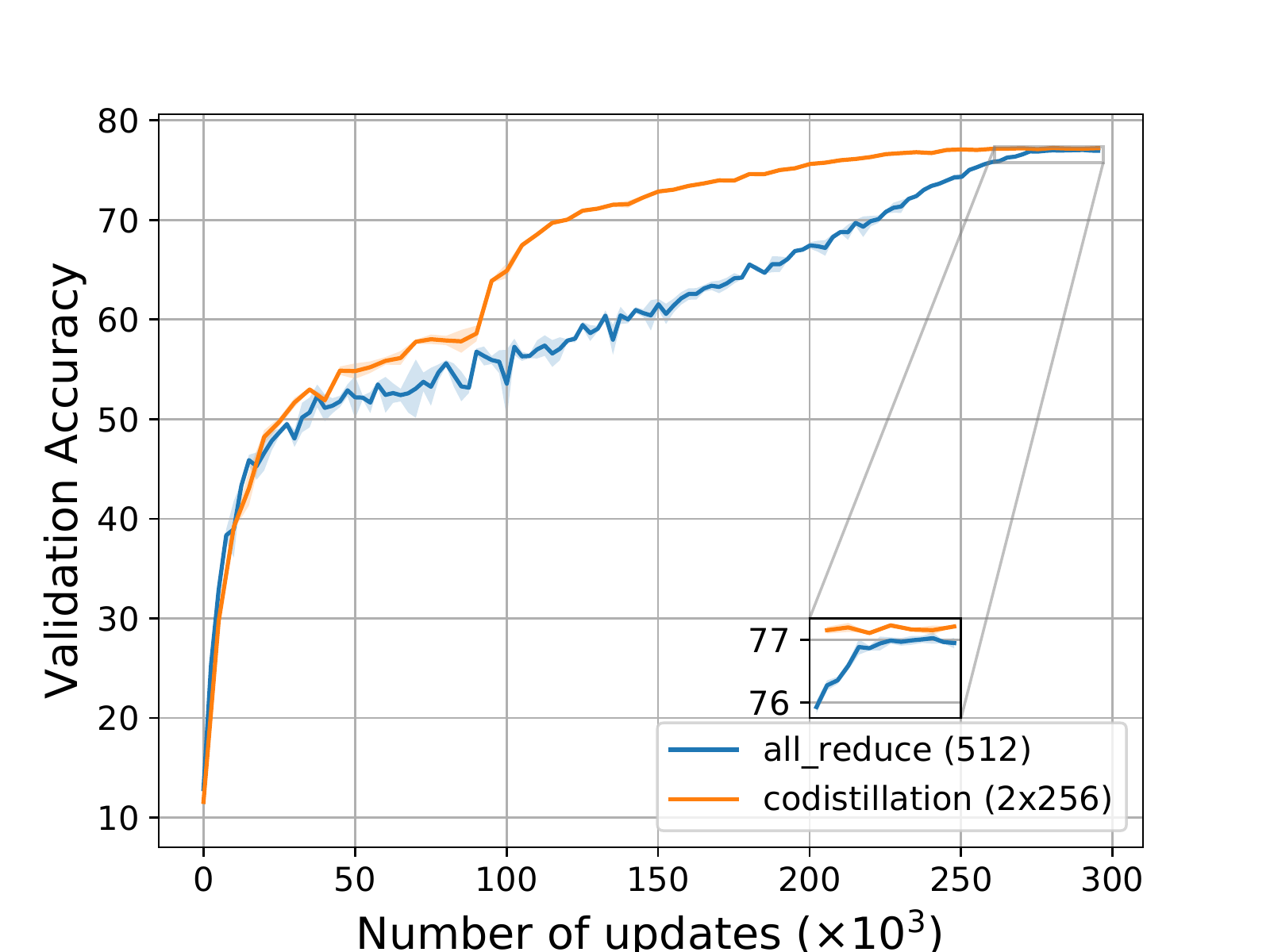}}
\subfigure[ResNeXt101]{\label{fig:compare_ar_with_codistillation_cosine_resnext101_validation_top_1_accuracy}\includegraphics[width=0.48\textwidth]{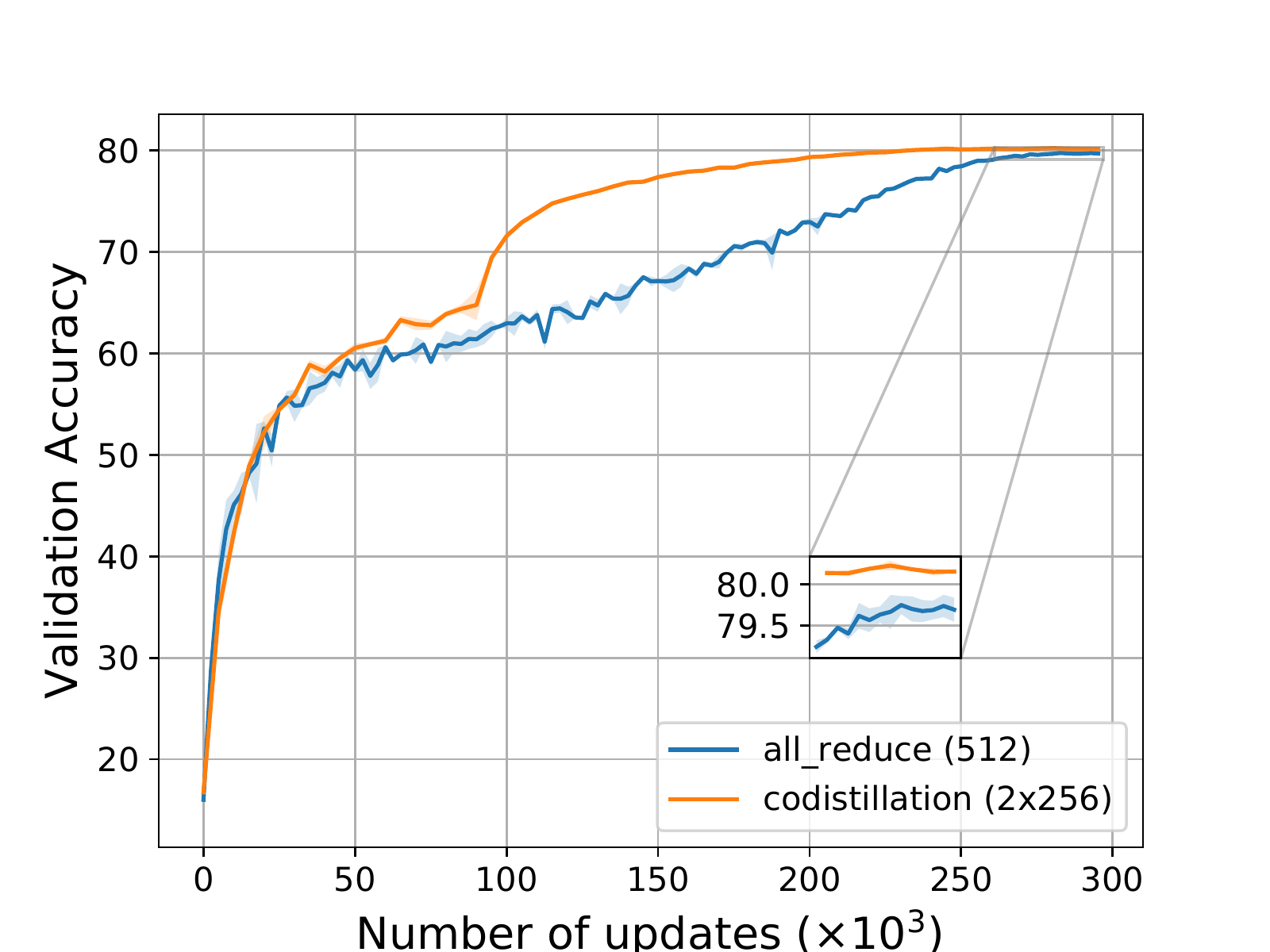}}
\caption{Comparing \allreduce and \codistillation for ResNet50 and ResNeXt101 models on the ImageNet dataset, using a cosine learning rate schedule~\citep{he2019bag_of_tricks_for_image_classification_with_convolutional_neural_networks}. We observe that the final validation performance for the two approaches is very close, confirming that \codistillation works consistently across different learning rate schedules.}
\label{fig:compare_ar_with_codistillation_cosine_validation_top_1_accuracy}
\end{figure}

Next, we demonstrate that the same strategy performs well when training other architectures and with other learning rate schedules. Specifically, we also train a ResNeXt101 model, and we use a half-cosine learning rate schedule~\cite{he2019bag_of_tricks_for_image_classification_with_convolutional_neural_networks} instead of the step-wise schedule. We use the same L2 regularization schedule in all codistillation experiments, with milestones at the iterations when the step-wise decays would have occurred in the tuned step-wise schedule. In Fig.~\ref{fig:compare_ar_with_codistillation_cosine_validation_top_1_accuracy}, we observe that the final validation performance for the two approaches is very close, confirming that \codistillation works consistently across different architectures and learning rate schedules; see Appendix~\ref{app_subsec:codistillation_vs_large_batch_training} for the corresponding plots of training loss and for additional related experiments.

\subsection{Neural Machine Translation}

\begin{figure}[t]
\centering
\includegraphics[width=0.45\textwidth]{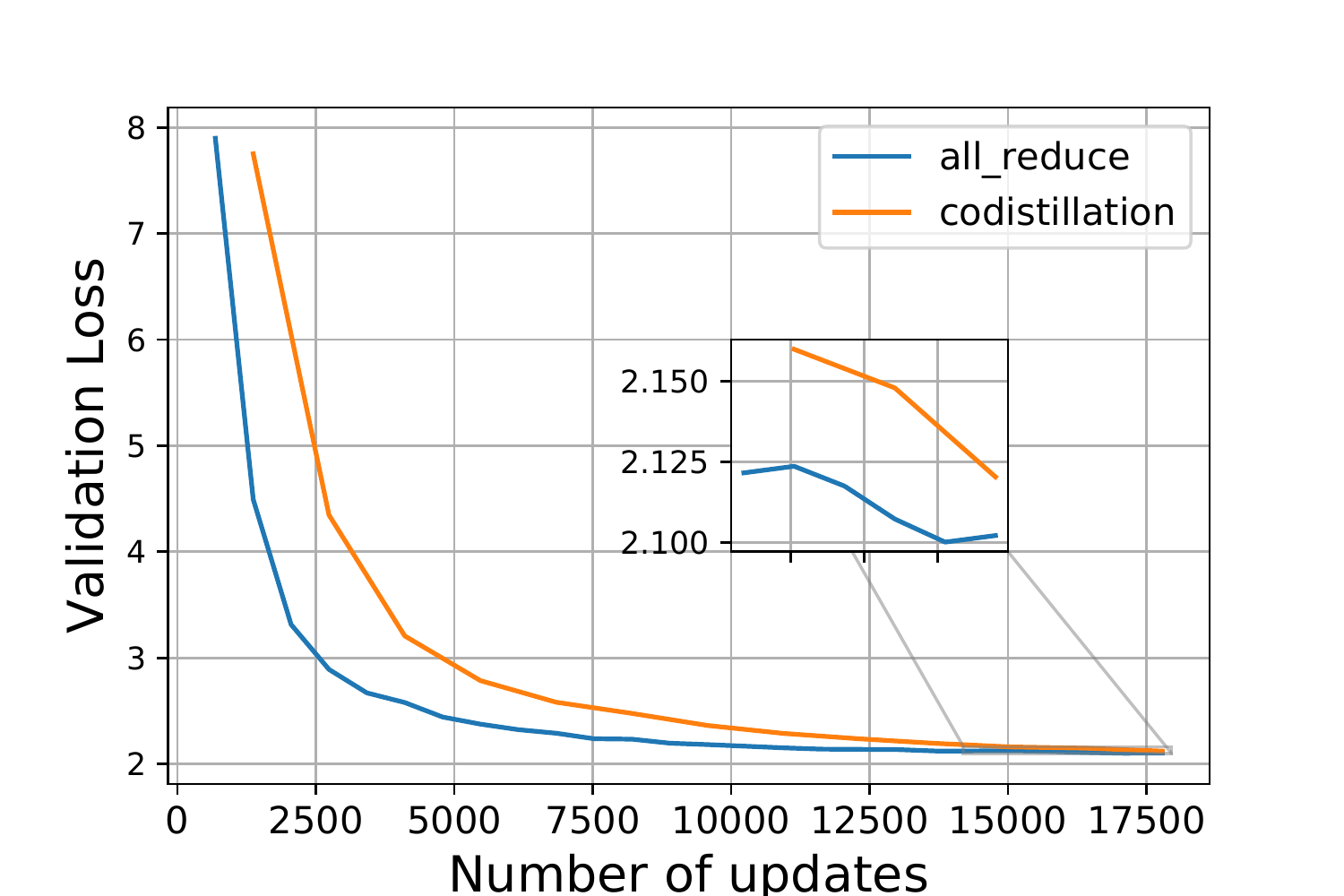}
\caption{Comparing \allreduce and \codistillation using ``big'' transformer model on WMT'16 En-De dataset. Models trained using \allreduce and codistillation both reach similar negative log-likelihood loss on the validation dataset.}
\label{fig:ar_vs_codistillation_transformer_wmt16_validation}
\end{figure}

Next we evaluate codistillation for training transformer models for machine translation. Specifically, we train the ``big'' transformer model~\citep{transformer_vaswani2017attention_is_all_you_need} (6 blocks in the encoder and decoder networks) on the WMT'16 En-De translation dataset, following the setup described in \cite{ott2018scaling_neural_machine_translation}. In Fig.~\ref{fig:ar_vs_codistillation_transformer_wmt16_validation}, we observe that the model trained with \codistillation reaches a similar validation loss as the model trained with \allreduce. For training such models it is more common to use label smoothing for regularization, rather than L2. Similar to the image classification experiments mentioned above, we decay the amount of label smoothing applied during training to obtain this result, as maintaining constant label smoothing also leads to overfitting and degraded performance; see Appendix~\ref{app:codistillation_works_beyond_vision} for details. This confirms that \codistillation with adjusted regularization schedule also extends to other workloads. The corresponding training plot is shown in Fig.~\ref{fig:ar_vs_codistillation_transformer_wmt16} in the appendix. %

\section{$n$-way Codistillation}

Next we investigate the performance of codistillation when scaling to $n > 2$. When increasing the number of models being codistilled, we may either hope to get the same accuracy after fewer updates (e.g., towards traditional linear scaling), or we may hope to achieve better accuracy after the same number of updates (e.g., via an ensembling effect). We explored both cases and found that: (i) with fewer updates (e.g., halving the number of updates per model when going from $n=2$ to $4$), codistillation achieves much worse accuracy(Fig.~\ref{fig:scale_codistillation_to_3_models_with_fixed_compute}); and (ii) even when performing the same number of updates after increasing $n$, we do not always observe noticeable gains.

\begin{figure}[h]
\centering     %
\subfigure[Training Loss with ResNet50]{\label{fig:codistillation_variants_train_loss}\includegraphics[width=0.48\textwidth]{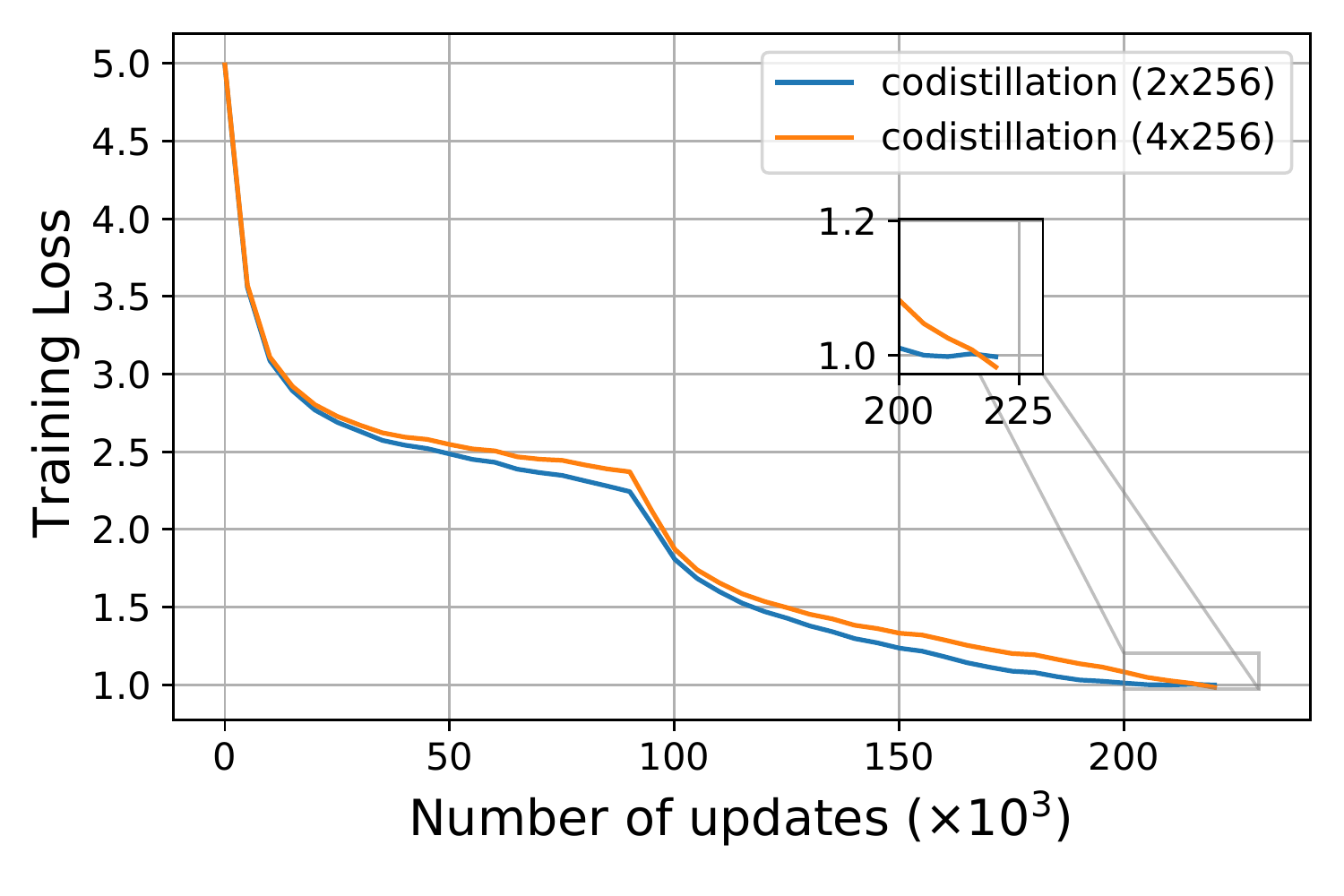}}
\subfigure[Validation Accuracy with ResNet50]{\label{fig:codistillation_variants_validation_accuracy}\includegraphics[width=0.48\textwidth]{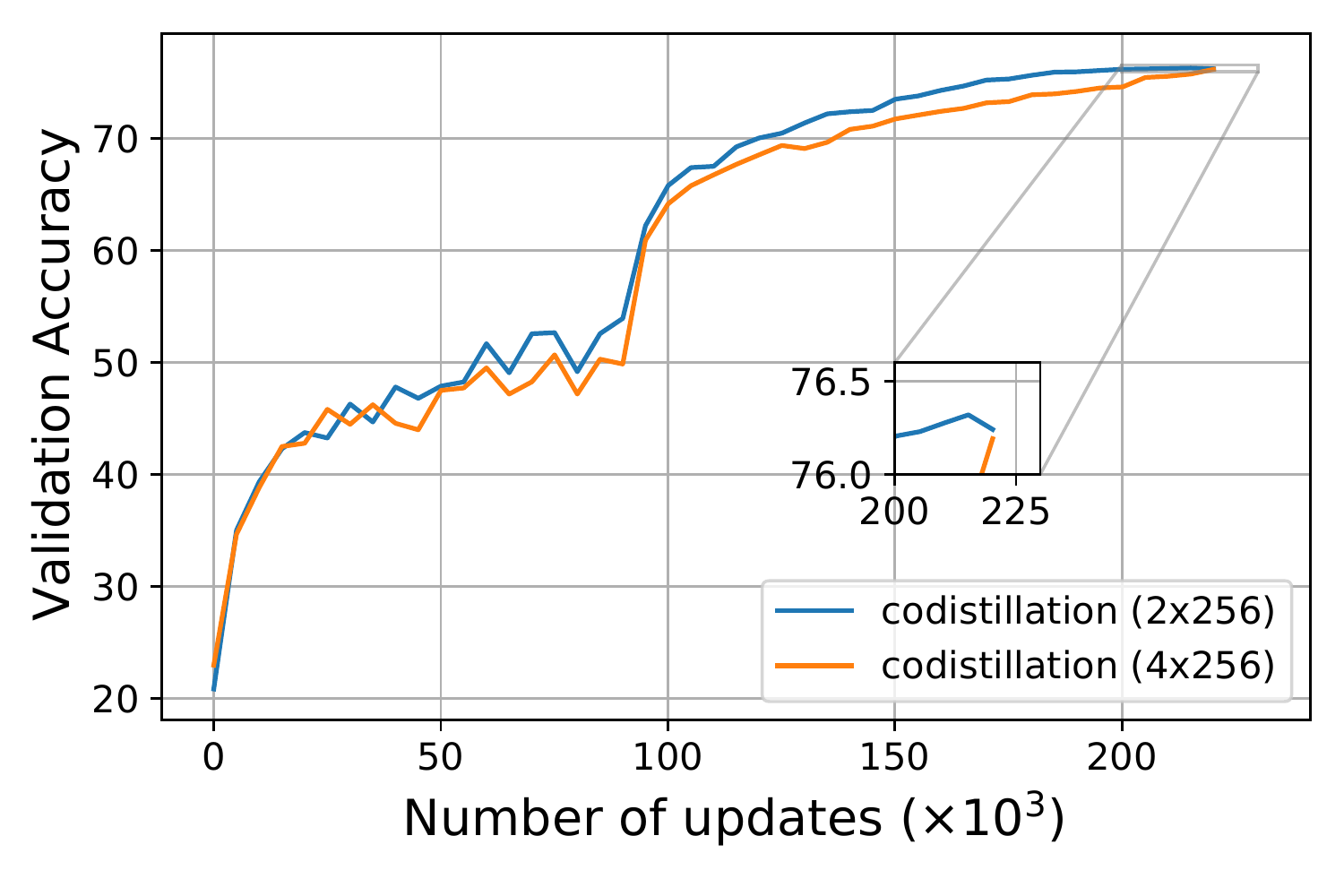}}
\caption{Comparing the training loss and validation accuracy (respectively) of 2-way and 4-way codistillation with ResNet-50 models on the ImageNet dataset. For this workload, the final model performance is the same for both values of $n$, suggesting that using codistillation with higher $n$ does not have any meaningful effect.}
\label{fig:scale_number_of_worker_groups}
\end{figure}

To illustrate the second point, we first consider an experiment where we compare $2$-way and $4$-way codistillation of transformer models for machine translation on the IWSLT dataset. In this case we observe some performance improvements (lower validation loss) when increasing $n$ (Table~\ref{tab:iwslt_n_way_codistillation} in the Appendix). In contrast, in Fig~\ref{fig:scale_number_of_worker_groups} we compare $2$-way and $4$-way codistillation of ResNet50 models on ImageNet and observe that there is no difference in performance when increasing $n$. This motivates us to better understand when we may expect to reap some benefits from using codistillation.

\subsection{A multi-view perspective}

Intuitively, from line~\ref{line:codist_model_update} in Algorithm~\ref{alg:codistillation}, one may expect benefits from increasing $n$ if the different models that are codistilling make sufficiently different predictions to provide informative training signal. In general, one may also expect to see diminishing returns when increasing $n$. \citet{allen-zhu_towards_2020} propose the so-called ``multi-view'' hypothesis in the context of traditional ensembling and distillation. Roughly speaking, the multi-view hypothesis suggests that the problem (dataset and model architecture) must possess certain structure making it possible for different model instances to make reasonable predictions based on distinct features (i.e., different ``views''). When different models do learn to make predictions by leveraging distinct features of the input, then one may expect that ensembling the models (or distilling them) will result in improved performance compared to any individual model. This is reminiscent of the boosting principle, and also may be loosely related to gradient diversity~\cite{yin2018gradient}.

Motivated by the multi-view hypothesis, we conduct the following experiment to confirm it also applies to $n$-way codistillation. To create a controlled setting where we know the problem contains multi-view structure, we use the CIFAR-10 dataset~\cite{krizhevsky2009learning} and begin with a pre-trained Wide-Resnet~(28x10) model, following the setup described in~\cite{allen-zhu_towards_2020}. We freeze the weights of the first bottleneck layer and reset the weights of all subsequent layers. Then we split the output channels of the first bottleneck layer into $8$ splits. The bottleneck layer of a Wide-Resnet~(28x10) model contains 160 channels, and so each split contains 20 $(= 160/8)$ channels. Now we consider $n$-way codistillation, with $n \in \{2, 4, 8\}$, and where each model is modified to only receive one of the splits after the first bottleneck layer. All the layers, other than the first frozen bottleneck layer, are trained. Because the splits come from a pre-trained model, we know that a model with access to all 160 channels after the full bottleneck is capable of achieving strong performance (top-1 accuracy of 94.92\%). When the different models being codistilled have the first bottleneck frozen to these pre-trained parameters and they use distinct splits, we also know that the models are receiving different views of the data.

\begin{figure}[t]
\centering
\includegraphics[width=0.5\textwidth]{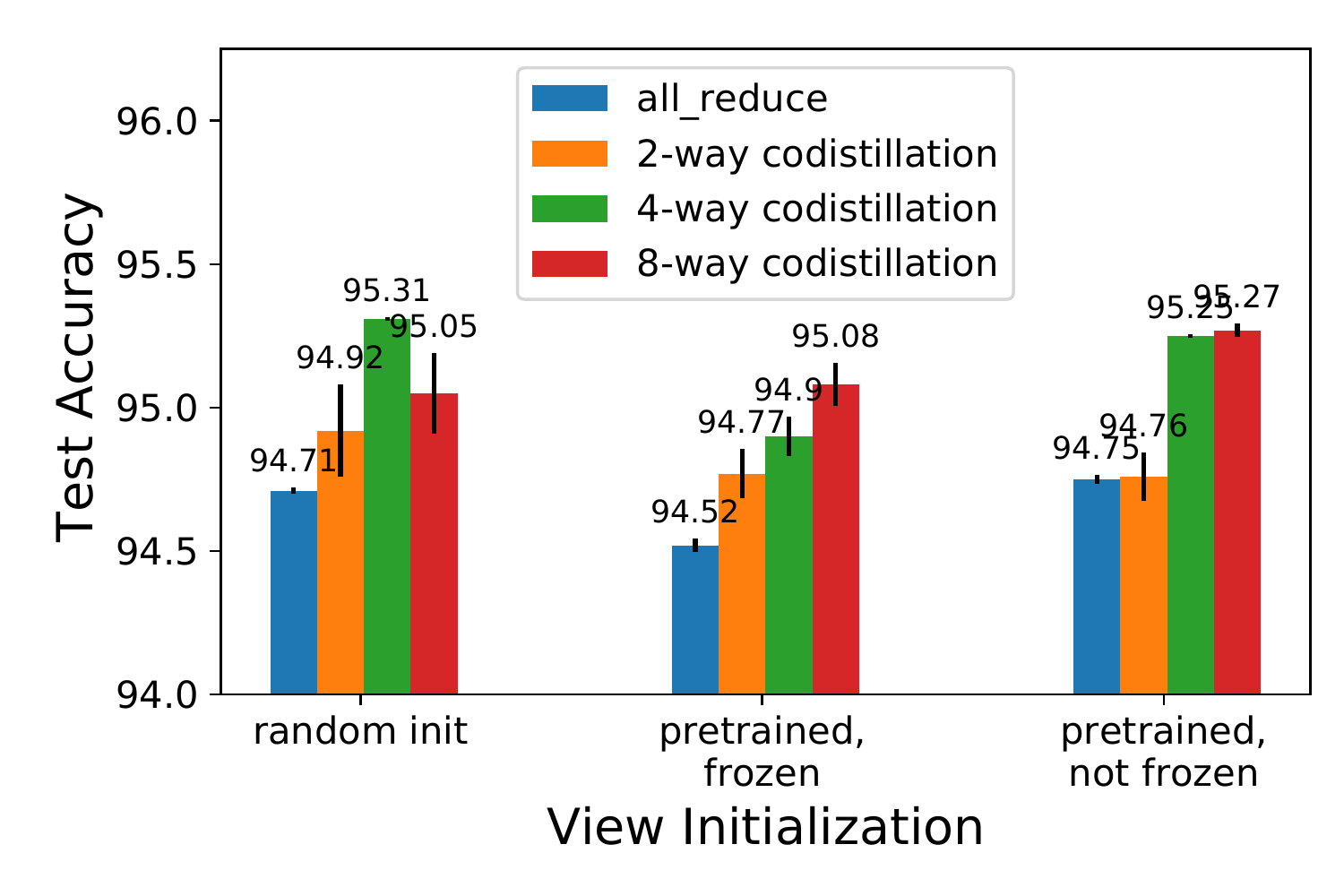}
\caption{Multi-view experiments towards understanding when increasing $n$ may benefit $n$-way codistillation. We compare \allreduce with $n$-way codistillation, for $n \in \{2, 4, 8\}$ in three different scenarios. All scenarios train a modified Wide-ResNet (28x10) model on CIFAR-10, where the architecture is modified so that the output of the first bottleneck layer has $1/8$ the number of channels compared to usual. When codistilling multiple models, we train all the models for the same number of steps (irrespective of $n$) and report the average top-1 accuracy across all models trained. The results shown are the mean (three seeds), and standard error(black lines).}
\label{fig:multi-view}
\end{figure}

In Fig.~\ref{fig:multi-view}, bars from the ``pretrained, frozen'' group show the results of this experiment. The performance with $n=1$ provides the average performance when training a modified model that only receives one of the $8$ splits, and provides a lower bound on performance. When we enforce the multi-view structure by keeping the weights of the first bottleneck layer frozen throughout training, we see that codistilling more models consistently improves performance when increasing $n$, confirming that one may expect codistillation to work well when the problem contains multi-view structure. The next question we ask is: is it sufficient to enforce the multi-view structure only at the start of the training or does it need to be enforced throughout training. Practically speaking, is it sufficient to start with the pretrained bottleneck layer or do we need to keep the layer frozen throughout training? In Fig.~\ref{fig:multi-view}, bars from the ``pretrained, not frozen'' group show the results of this experiment. We observe a performance improvement from $2$-way codistillation to $4$-way codistillation but beyond that, the performance increase is negligible. This observation suggests that the multi-view structure needs to be maintained throughout training. Finally, we want to study the effect of increasing $n$ when we train the model from scratch. Bars from the ``random init'' group show the results of this setting. All models are trained with the same one split of the data. We observe that while we are able to improve performance by increasing $n$ to 4, performance degrades as we increase $n$ to 8. These results suggest that as we increase $n$, codistillation helps to improve performance as long as the ``multi-view'' hypothesis holds for the problem setup. Pretraining (followed by splitting the features) can be a good strategy to get data with ``multiple views'', but just initializing the training with ``multi-view'' data is likely not sufficient when scaling codistillation to multiple models.

\subsection{Multiple views from multiple architectures}

The experiment discussed in the previous subsection suggests that when multi-view structure is present in the problem (a combined feature of the training dataset and model architecture), when codistilling multiple copies of the same architecture it may be possible for the codistilled models to benefit from this structure. One approach to potentially injecting multi-view structure could be to codistill different architectures. We test this hypothesis by codistilling between ResNet18, ResNet50 and ResNext101 models on the ImageNet dataset. As shown in Fig.~\ref{fig:3_way_codistillation_between_different_architectures_3_way} (in Appendix), the performance of the ResNet50 model improves compared to the case where the model is trained only with ResNet50 models (either in \allreduce or $2-$way/$4-$way codisitllation). The gain in performance of ResNet50 could happen either because: (i) we are using different architectures, (ii) we are codistilling $>2$ models or both (i) and (ii). We design another experiment to understand why codistillation performs well in this case. Specifically, we perform $2-$way codistillation between a ResNet50 and ResNext101 model. If we do not see any performance improvement, then we can conclude that the performance benefit is because of (ii) i.e codistilling between more than $>2$ models. In Fig.~\ref{fig:3_way_codistillation_between_different_architectures_2_way_vs_3_way_different_architectures}, we report that the ResNet50 model achieves a much better performance with this new $2-$way setup as it would with the earlier $3-$way setup. This observation suggests that the earlier performance gain was due to codistillation with a larger model and not because of codistillation between $>2$ models. Infact, codistilling with a smaller ResNet-18 model (while also codistilling with the larger ResNext-101 model) could likely over-regularize the model, as seems to be the case here. This observation suggests that while codistilling with a larger model is better in practice than codistilling with a smaller model, it may not help to scale codistillation to $n>2$ models, thus leaving it an open problem.

\section{Trade-offs With Codistillation}
\label{sec:tradeoffs_with_codistillation}
So far, we have focused on evaluating codistillation's feasibility for distributed training, understanding how it works, how it affects the training dynamics, if it can be scaled by adding more workers (per model) or by increasing the number of codistilled models. In this section, we highlight the different trade-offs of codistillation and if/when it can be used as an alternative to \allreduce. While we mention some \textit{limitations} of codistillation, we do not view them as \textit{failures} of codistillation. We believe that these current limitations are opportunities for improving our understanding of codistillation and making it more viable for distributed training.

\begin{enumerate}
    \item Codistillation may not completely replace \allreduce. Both our work and previous works~\citep{anil2018large} used codistillation along with \allreduce (i.e. when codistilling multiple models, each model is trained using \allreduce). We showed that it is difficult to scale \codistillation to more than $2$ models. This observation suggests that using \codistillation to fully replace \allreduce, for example by replacing \allreduce between 16 devices by codistillation between 16 models (each model using one device), is not likely to work in practice.
    \item While $2-$way codistillation can be competitive with \allreduce setup (Fig.~\ref{fig:compare_ar_with_codistillation_cosine_validation_top_1_accuracy}), and can be scaled to work with a large number of workers (Fig.~\ref{fig:compare_codistillation_for_differernt_batch_sizes_resnet50_validation_top_1_accuracy}), codistillation can also introduce some unintended effects in the training dynamics (like regularization effect).
    \item Codistillation can reduce the communication between the models (Fig.~\ref{fig:comparing_bits}). However, it may not necessarily lead to faster training (in terms of wall clock time). Any performance benefit (in terms of wall clock time) depends on additional parameters such as the size of the model (large enough models are likely to benefit more) and bandwidth of the communication channel (more benefits with slower bandwidth). 
    \item Codistillation introduces an additional hyper-parameter $\alpha$. However, this may not be a major concern as it is relatively easy to tune.
    \item Scaling codistillation to $n>2$ models presents several challenges related to problem-specific characteristics of the model architecture and the dataset (Fig.~\ref{fig:multi-view}).
    \item Codistillation provides a weaker yet more flexible synchronization mechanism as compared to \allreduce, thus enabling many interesting applications. For example, codistillation can be used to increase the performance of a given model by codisitilling it with a larger model. This is similar to the benefit we obtain by creating an ensemble of different models but has the additional advantage that only one model is needed during inference. 
\end{enumerate}

\section{Conclusion}
\label{sec:conclusion}

In this work, we demonstrate for the first time (to the best of our knowledge), that models trained with \codistillation can perform as well as models trained with traditional parallel SGD methods (while using comparable computational resources). Complementing existing works, we study how codistillation affects the training dynamics. We highlight the regularizing effect of codistillation and show that accounting for this effect is essential for obtaining good performance using \codistillation. We show that while codistillation can be scaled by using more devices per model (similar to how parallel SGD methods are scaled), it is much harder to scale codistillation by increasing the number of codistilling models for many workloads. 

There are several exciting and potentially impactful directions for extending our understanding of \codistillation. While we demonstrated some examples where codistillation can bring performance improvements with $>2$ models, there are other examples where $(n>2)$-way codistillation is no better than $2-$way codistillation. It is not clear if something can be changed in the training setup (e.g., changing how models are initialized) so that $n-$way codistillation can provide improved performance more consistently. Codistillation reduces the number of bits to be communicated while training the model. It will be interesting to study the full impact of this reduction in the case of training on low-resource devices. The loosely synchronized nature of \codistillation opens up interesting possibilities such as codistillation between different architectures and/or models trained on different datasets. An interesting follow up would be study how this flexibility can be leveraged for improving generalization (or robustness) properties of the models.

\paragraph{Broader Impact.} Large-scale machine learning models have several important social benefits like improved language translation (making information accessible to more people) and hateful content detection at scale. At the same time, training such models introduces several challenges - from the availability of compute resources and data to utilizing these resources efficiently and effectively. Data parallel training is the dominant technique for scaling neural network training. Large-scale distributed training also has environmental implications~\cite{henderson2020towards}. In this work, we study a complementary approach called codistillation that can be used with data parallel training techniques to speed up training by reducing the communication overhead when training models. We note that there are several other challenges (like fairness, bias etc) that arise when using large-scale models in practice. While it is very important to study and understand these challenges, addressing these challenges will require developing complementary techniques that are beyond the scope of codistillation.

\hypersetup{
    linkcolor=blue,
    filecolor=magenta,      
    urlcolor=cyan,
    }

\bibliographystyle{abbrvnat}
\bibliography{codistillation}

\begin{thebibliography}{27}
\providecommand{\natexlab}[1]{#1}
\providecommand{\url}[1]{\texttt{#1}}
\expandafter\ifx\csname urlstyle\endcsname\relax
  \providecommand{\doi}[1]{doi: #1}\else
  \providecommand{\doi}{doi: \begingroup \urlstyle{rm}\Url}\fi

\bibitem[Alistarh et~al.(2017)Alistarh, Grubic, Li, Tomioka, and
  Vojnovic]{alistarh2017qsgd}
D.~Alistarh, D.~Grubic, J.~Li, R.~Tomioka, and M.~Vojnovic.
\newblock Qsgd: Communication-efficient sgd via gradient quantization and
  encoding.
\newblock In \emph{Advances in Neural Information Processing Systems}, pages
  1709--1720, 2017.

\bibitem[Allen-Zhu and Li(2020)]{allen-zhu_towards_2020}
Z.~Allen-Zhu and Y.~Li.
\newblock Towards {Understanding} {Ensemble}, {Knowledge} {Distillation} and
  {Self}-{Distillation} in {Deep} {Learning}.
\newblock \emph{arXiv preprint}, Dec. 2020.
\newblock URL \url{https://arxiv.org/abs/2012.09816v1}.

\bibitem[Anil et~al.(2018)Anil, Pereyra, Passos, Ormandi, Dahl, and
  Hinton]{anil2018large}
R.~Anil, G.~Pereyra, A.~Passos, R.~Ormandi, G.~E. Dahl, and G.~E. Hinton.
\newblock Large scale distributed neural network training through online
  distillation.
\newblock In \emph{International Conference on Learning Representations
  (ICLR)}, 2018.

\bibitem[Assran et~al.(2019)Assran, Loizou, Ballas, and
  Rabbat]{sgp_pmlr-v97-assran19a}
M.~Assran, N.~Loizou, N.~Ballas, and M.~Rabbat.
\newblock Stochastic gradient push for distributed deep learning.
\newblock In K.~Chaudhuri and R.~Salakhutdinov, editors, \emph{International
  Conference on Machine Learning (ICML)}, volume~97 of \emph{Proceedings of
  Machine Learning Research}, pages 344--353, Long Beach, California, USA,
  09--15 Jun 2019. PMLR.

\bibitem[Brown et~al.(2020)Brown, Mann, Ryder, Subbiah, Kaplan, Dhariwal,
  Neelakantan, Shyam, Sastry, Askell,
  et~al.]{gpt3_brown2020language_models_are_few_shot_learners}
T.~B. Brown, B.~Mann, N.~Ryder, M.~Subbiah, J.~Kaplan, P.~Dhariwal,
  A.~Neelakantan, P.~Shyam, G.~Sastry, A.~Askell, et~al.
\newblock Language models are few-shot learners.
\newblock \emph{arXiv preprint arXiv:2005.14165}, 2020.

\bibitem[de~la Chimie(2010)]{de2010international}
M.~de~la Chimie.
\newblock International workshop on spoken language translation.
\newblock 2010.

\bibitem[Devlin et~al.(2018)Devlin, Chang, Lee, and Toutanova]{devlin2018bert}
J.~Devlin, M.-W. Chang, K.~Lee, and K.~Toutanova.
\newblock Bert: Pre-training of deep bidirectional transformers for language
  understanding.
\newblock \emph{arXiv preprint arXiv:1810.04805}, 2018.

\bibitem[Goodfellow et~al.(2016)Goodfellow, Bengio, Courville, and
  Bengio]{goodfellow2016deep}
I.~Goodfellow, Y.~Bengio, A.~Courville, and Y.~Bengio.
\newblock \emph{Deep learning}, volume~1.
\newblock MIT press Cambridge, 2016.

\bibitem[Goyal et~al.(2017)Goyal, Doll{\'a}r, Girshick, Noordhuis, Wesolowski,
  Kyrola, Tulloch, Jia, and He]{goyal2017imagenet_in_an_hour}
P.~Goyal, P.~Doll{\'a}r, R.~Girshick, P.~Noordhuis, L.~Wesolowski, A.~Kyrola,
  A.~Tulloch, Y.~Jia, and K.~He.
\newblock Accurate, large minibatch sgd: Training imagenet in 1 hour.
\newblock \emph{arXiv preprint arXiv:1706.02677}, 2017.

\bibitem[He et~al.(2016)He, Zhang, Ren, and
  Sun]{resnet_he2016deep_residual_learning_for_image_recognition}
K.~He, X.~Zhang, S.~Ren, and J.~Sun.
\newblock Deep residual learning for image recognition.
\newblock In \emph{Proceedings of the IEEE conference on computer vision and
  pattern recognition}, pages 770--778, 2016.

\bibitem[He et~al.(2019)He, Zhang, Zhang, Zhang, Xie, and
  Li]{he2019bag_of_tricks_for_image_classification_with_convolutional_neural_networks}
T.~He, Z.~Zhang, H.~Zhang, Z.~Zhang, J.~Xie, and M.~Li.
\newblock Bag of tricks for image classification with convolutional neural
  networks.
\newblock In \emph{Proceedings of the IEEE Conference on Computer Vision and
  Pattern Recognition}, pages 558--567, 2019.

\bibitem[Henderson et~al.(2020)Henderson, Hu, Romoff, Brunskill, Jurafsky, and
  Pineau]{henderson2020towards}
P.~Henderson, J.~Hu, J.~Romoff, E.~Brunskill, D.~Jurafsky, and J.~Pineau.
\newblock Towards the systematic reporting of the energy and carbon footprints
  of machine learning.
\newblock \emph{Journal of Machine Learning Research}, 21:\penalty0 1--43,
  2020.

\bibitem[Hinton et~al.(2014)Hinton, Vinyals, and
  Dean]{hinton2014distilling_the_knowledge_in_a_neural_network}
G.~Hinton, O.~Vinyals, and J.~Dean.
\newblock Distilling the knowledge in a neural network.
\newblock In \emph{NeurIPS Deep Learning Workshop}, 2014.

\bibitem[Huang et~al.(2019)Huang, Cheng, Bapna, Firat, Chen, Chen, Lee, Ngiam,
  Le, Wu, et~al.]{huang2019gpipe}
Y.~Huang, Y.~Cheng, A.~Bapna, O.~Firat, D.~Chen, M.~Chen, H.~Lee, J.~Ngiam,
  Q.~V. Le, Y.~Wu, et~al.
\newblock Gpipe: Efficient training of giant neural networks using pipeline
  parallelism.
\newblock In \emph{Advances in neural information processing systems}, pages
  103--112, 2019.

\bibitem[Johnson et~al.(2020)Johnson, Agrawal, Gu, and
  Guestrin]{johnson2020adascale}
T.~B. Johnson, P.~Agrawal, H.~Gu, and C.~Guestrin.
\newblock {AdaScale SGD}: A user-friendly algorithm for distributed training.
\newblock In \emph{International Conference on Machine Learning}, 2020.

\bibitem[Kaplan et~al.(2020)Kaplan, McCandlish, Henighan, Brown, Chess, Child,
  Gray, Radford, Wu, and
  Amodei]{kaplan2020scaling_laws_for_neural_language_models}
J.~Kaplan, S.~McCandlish, T.~Henighan, T.~B. Brown, B.~Chess, R.~Child,
  S.~Gray, A.~Radford, J.~Wu, and D.~Amodei.
\newblock Scaling laws for neural language models.
\newblock \emph{arXiv preprint arXiv:2001.08361}, 2020.

\bibitem[Krizhevsky et~al.(2009)Krizhevsky, Hinton,
  et~al.]{krizhevsky2009learning}
A.~Krizhevsky, G.~Hinton, et~al.
\newblock Learning multiple layers of features from tiny images.
\newblock \emph{arXiv preprint}, 2009.

\bibitem[Lepikhin et~al.(2020)Lepikhin, Lee, Xu, Chen, Firat, Huang, Krikun,
  Shazeer, and Chen]{lepikhin2020gshard}
D.~Lepikhin, H.~Lee, Y.~Xu, D.~Chen, O.~Firat, Y.~Huang, M.~Krikun, N.~Shazeer,
  and Z.~Chen.
\newblock Gshard: Scaling giant models with conditional computation and
  automatic sharding.
\newblock \emph{arXiv preprint arXiv:2006.16668}, 2020.

\bibitem[Ott et~al.(2018)Ott, Edunov, Grangier, and
  Auli]{ott2018scaling_neural_machine_translation}
M.~Ott, S.~Edunov, D.~Grangier, and M.~Auli.
\newblock Scaling neural machine translation.
\newblock \emph{arXiv preprint arXiv:1806.00187}, 2018.

\bibitem[Ott et~al.(2019)Ott, Edunov, Baevski, Fan, Gross, Ng, Grangier, and
  Auli]{ott2019fairseq}
M.~Ott, S.~Edunov, A.~Baevski, A.~Fan, S.~Gross, N.~Ng, D.~Grangier, and
  M.~Auli.
\newblock fairseq: A fast, extensible toolkit for sequence modeling.
\newblock In \emph{Proceedings of NAACL-HLT 2019: Demonstrations}, 2019.

\bibitem[Paszke et~al.(2017)Paszke, Gross, Chintala, Chanan, Yang, DeVito, Lin,
  Desmaison, Antiga, and Lerer]{paszke2017pytorch}
A.~Paszke, S.~Gross, S.~Chintala, G.~Chanan, E.~Yang, Z.~DeVito, Z.~Lin,
  A.~Desmaison, L.~Antiga, and A.~Lerer.
\newblock Automatic differentiation in pytorch.
\newblock \emph{arXiv preprint}, 2017.

\bibitem[Russakovsky et~al.(2015)Russakovsky, Deng, Su, Krause, Satheesh, Ma,
  Huang, Karpathy, Khosla, Bernstein,
  et~al.]{russakovsky2015imagenet_large_scale_visual_recognition_challenge}
O.~Russakovsky, J.~Deng, H.~Su, J.~Krause, S.~Satheesh, S.~Ma, Z.~Huang,
  A.~Karpathy, A.~Khosla, M.~Bernstein, et~al.
\newblock Imagenet large scale visual recognition challenge.
\newblock \emph{International Journal of Computer Vision}, 115\penalty0
  (3):\penalty0 211--252, 2015.

\bibitem[Shoeybi et~al.(2019)Shoeybi, Patwary, Puri, LeGresley, Casper, and
  Catanzaro]{shoeybi2019megatron}
M.~Shoeybi, M.~Patwary, R.~Puri, P.~LeGresley, J.~Casper, and B.~Catanzaro.
\newblock Megatron-{LM}: Training multi-billion parameter language models using
  {GPU} model parallelism.
\newblock \emph{arXiv preprint arXiv:1909.08053}, 2019.

\bibitem[Stich(2018)]{stich2018local_sgd_converges_fast_and_communicates_little}
S.~U. Stich.
\newblock Local {SGD} converges fast and communicates little.
\newblock \emph{arXiv preprint arXiv:1805.09767}, 2018.

\bibitem[Vaswani et~al.(2017)Vaswani, Shazeer, Parmar, Uszkoreit, Jones, Gomez,
  Kaiser, and Polosukhin]{transformer_vaswani2017attention_is_all_you_need}
A.~Vaswani, N.~Shazeer, N.~Parmar, J.~Uszkoreit, L.~Jones, A.~N. Gomez,
  {\L}.~Kaiser, and I.~Polosukhin.
\newblock Attention is all you need.
\newblock In \emph{Advances in neural information processing systems}, pages
  5998--6008, 2017.

\bibitem[Yin et~al.(2018)Yin, Pananjady, Lam, Papailiopoulos, Ramchandran, and
  Bartlett]{yin2018gradient}
D.~Yin, A.~Pananjady, M.~Lam, D.~Papailiopoulos, K.~Ramchandran, and
  P.~Bartlett.
\newblock Gradient diversity: A key ingredient for scalable distributed
  learning.
\newblock In \emph{International Conference on Artificial Intelligence and
  Statistics (AISTATS)}, pages 1998--2007, 2018.

\bibitem[Zhang et~al.(2018)Zhang, Xiang, Hospedales, and Lu]{zhang2018deep}
Y.~Zhang, T.~Xiang, T.~M. Hospedales, and H.~Lu.
\newblock Deep mutual learning.
\newblock In \emph{IEEE Conf. on Computer Vision and Pattern Recognition
  (CVPR)}, pages 4320--4328, 2018.

\end{thebibliography}

\section*{Checklist}

\begin{enumerate}

\item For all authors...
\begin{enumerate}
  \item Do the main claims made in the abstract and introduction accurately reflect the paper's contributions and scope?
    \answerYes{}
  \item Did you describe the limitations of your work?
    \answerYes{In Section~\ref{sec:tradeoffs_with_codistillation}}
  \item Did you discuss any potential negative societal impacts of your work?
    \answerYes{In Section~\ref{sec:conclusion}}
  \item Have you read the ethics review guidelines and ensured that your paper conforms to them?
    \answerYes{}
\end{enumerate}

\item If you are including theoretical results...
\begin{enumerate}
  \item Did you state the full set of assumptions of all theoretical results?
    \answerNA{}
	\item Did you include complete proofs of all theoretical results?
    \answerNA{}
\end{enumerate}

\item If you ran experiments...
\begin{enumerate}
  \item Did you include the code, data, and instructions needed to reproduce the main experimental results (either in the supplemental material or as a URL)?
    \answerYes{}
  \item Did you specify all the training details (e.g., data splits, hyperparameters, how they were chosen)?
    \answerYes{Section~\ref{subsec:implementation_details}}
	\item Did you report error bars (e.g., with respect to the random seed after running experiments multiple times)?
    \answerYes{Shaded region in the figures.}
	\item Did you include the total amount of compute and the type of resources used (e.g., type of GPUs, internal cluster, or cloud provider)?
    \answerYes{Section~\ref{subsec:implementation_details}}
\end{enumerate}

\item If you are using existing assets (e.g., code, data, models) or curating/releasing new assets...
\begin{enumerate}
  \item If your work uses existing assets, did you cite the creators?
    \answerYes{Section~\ref{subsec:implementation_details}}
  \item Did you mention the license of the assets?
    \answerYes{Section~\ref{subsec:implementation_details}}
  \item Did you include any new assets either in the supplemental material or as a URL?
    \answerYes{}
  \item Did you discuss whether and how consent was obtained from people whose data you're using/curating?
    \answerYes{Section~\ref{subsec:dataset_details}}
  \item Did you discuss whether the data you are using/curating contains personally identifiable information or offensive content?
    \answerYes{Section~\ref{subsec:dataset_details}}
\end{enumerate}

\item If you used crowdsourcing or conducted research with human subjects...
\begin{enumerate}
  \item Did you include the full text of instructions given to participants and screenshots, if applicable?
    \answerNA{}
  \item Did you describe any potential participant risks, with links to Institutional Review Board (IRB) approvals, if applicable?
    \answerNA{}
  \item Did you include the estimated hourly wage paid to participants and the total amount spent on participant compensation?
    \answerNA{}
\end{enumerate}

\end{enumerate}

\clearpage 

\appendix

\begin{center}
    \Large \bf A Closer Look at Codistillation \\ Supplementary Material
\end{center}

\section{Appendix}

\subsection{Implementation Details}
\label{app:implementation_options}

\subsection{Dataset Details}
\label{subsec:dataset_details}

We use the following datasets:

\begin{enumerate}
    \item ImageNet Dataset\footnote{https://www.image-net.org/index.php} is an image database organized as per the WordNet hierarchy. The most commonly used subset of ImageNet is the 1000-class subset~\cite{russakovsky2015imagenet_large_scale_visual_recognition_challenge} and is the subset used om this work as well. The full ImageNet dataset has some classes that may lead to problematic beahviors in the trained models. In 2020, about 2702 synsets in the ``person'' subtree were removed to account for this possibility\footnote{https://www.image-net.org/update-mar-11-2021.php}. However this update does not affect the 1000 class subset that we are using. 
    \item CIFAR-10\footnote{https://www.cs.toronto.edu/~kriz/cifar.html} is a labeled subset of the 80 million tiny images dataset~\citep{krizhevsky2009learning}. It consists of 60000 32x32 colour images in 10 classes, with 6000 images per class. There are 50000 training images and 10000 test images.
    \item IWSLT'14 German to English Dataset ~\cite{de2010international} has about 153k/7k/7k parallel sentences as training/dev/test. Vocabulary size is 32K symbols.
    \item WMT'16 English to German Dataset\footnote{http://www.statmt.org/wmt16/translation-task.html} has about 4.56M/3K/2.6K parallel sentences as training/dev/test. The training data is combined from Europarl v7, Common Crawl, and News Commentary v11. Vocabulary size is 32K symbols.
\end{enumerate}

\subsection{Implementation Details}
\label{subsec:implementation_details}

The code for all experiments is implemented using PyTorch~\citep{paszke2017pytorch}\footnote{https://pytorch.org/}. All experiments are run for 3 seeds and plots report the average over all seeds. The \codistillation loss is the mean squared error between the logits of the two models (without centering them, as we found in preliminary experiments that centering yielded similar results). 

\paragraph{License}

\begin{enumerate}
    \item ImageNet: \href{https://www.image-net.org/download.php}{https://www.image-net.org/download.php}
    \item CIFAR: \href{https://www.cs.toronto.edu/~kriz/cifar.html}{https://www.cs.toronto.edu/~kriz/cifar.html}
    \item IWSLT-14: \href{https://sites.google.com/site/iwsltevaluation2014/data-provided}{https://sites.google.com/site/iwsltevaluation2014/data-provided}
    \item WMT'16 Englis to German: \href{http://www.statmt.org/wmt16/translation-task.html}{http://www.statmt.org/wmt16/translation-task.html}
    \item PyTorch: \href{https://github.com/pytorch/pytorch/blob/master/LICENSE}{https://github.com/pytorch/pytorch/blob/master/LICENSE}
    \item PyTorch-Examples: BSD 3-Clause \href{https://github.com/pytorch/examples/blob/master/LICENSE}{https://github.com/pytorch/examples/blob/master/LICENSE}
    \item FairSeq: MIT License \href{https://github.com/pytorch/fairseq/blob/master/LICENSE}{https://github.com/pytorch/fairseq/blob/master/LICENSE}
    \item OmegaConf: BSD 3-Clause \href{https://github.com/omry/omegaconf/blob/master/LICENSE}{https://github.com/omry/omegaconf/blob/master/LICENSE}
\end{enumerate}

\paragraph{ImageNet Experiments}

The experiments involving the ImageNet dataset are based on the setup proposed in \cite{goyal2017imagenet_in_an_hour}. We use a constant value for $\alpha$ i.e. $\alpha^k = 1$ for all $k$. All the experiments are run on Volta100 16 GB GPU with a batch size of 32 per GPU (unless specified otherwise).

\paragraph{Neural Machine Translation Experiments}

The experiments for translation tasks use the FairSeq library~\citep{ott2019fairseq}\footnote{https://github.com/pytorch/fairseq/releases/tag/v0.9.0}. These experiments are based on the setup proposed in \cite{ott2018scaling_neural_machine_translation}. We increase $\alpha^k$ by a factor $\gamma = 1.1$ every epoch. When training the translation models with \codistillation, we explicitly reduce the amount of regularization (to account for regularization added by \codistillation) by removing the label smoothing loss. All the experiments are run on Volta100 32 GB GPU with 3854 tokens per GPU.

\begin{wrapfigure}{f}{0.5\textwidth}
    \begin{center}
        \includegraphics[width=0.5\textwidth]{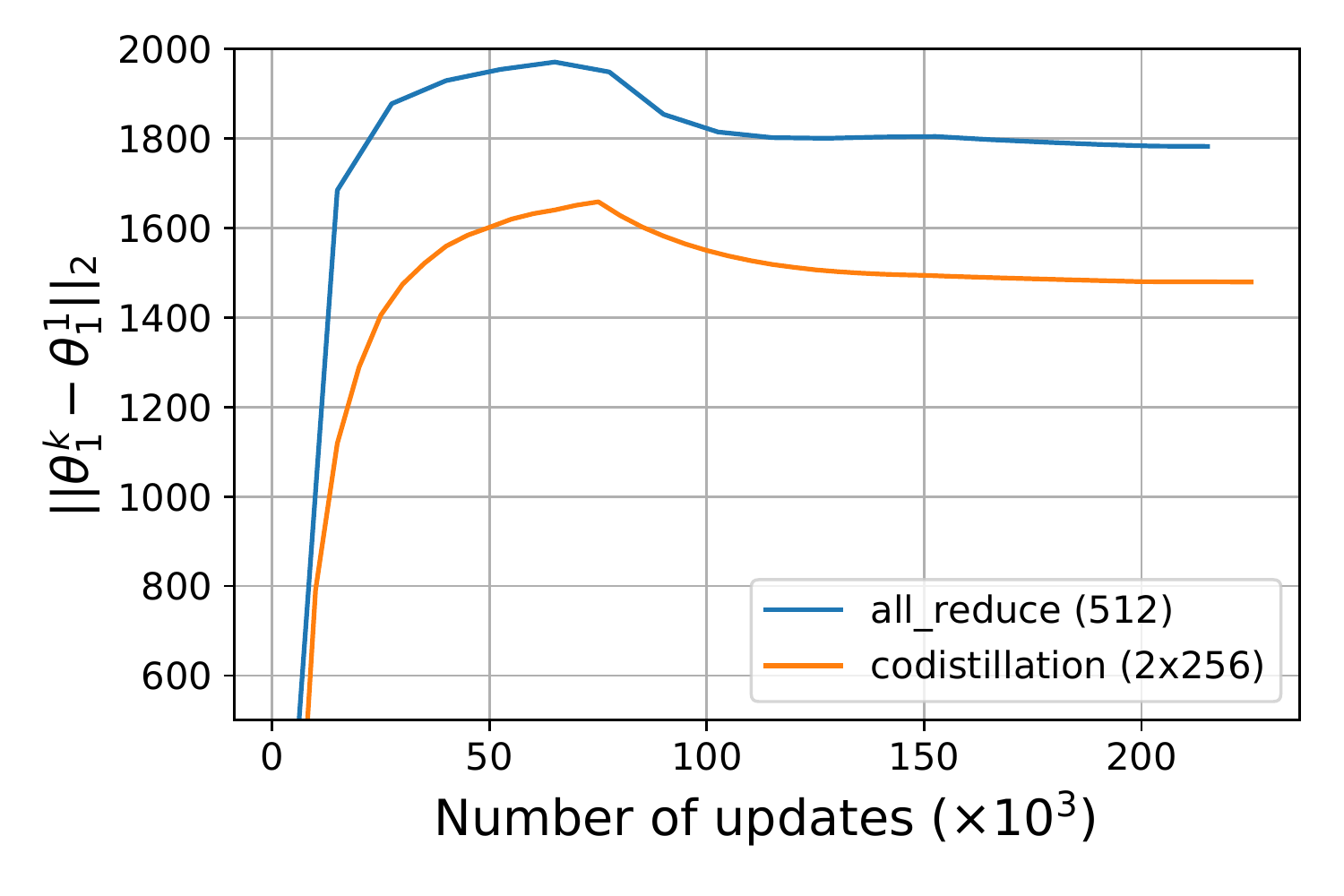}
    \end{center}
\caption{We examine how the difference in parameters evolves from initialization over the course of training. When training with \codistillation, the parameters remain closer to their initial values, thus demonstrating the regularizing effect of codistillation.}
\label{fig:compare_ar_with_codistillation_param_change}
\vspace{-10pt}
\end{wrapfigure}

\subsection{Additonal Results When Using Codistillaton For Distributed Training For Vision}
\label{app_subsec:codistillation_vs_large_batch_training}

\subsubsection{Demonstrating Regularization Effect}

In Fig.~\ref{fig:compare_ar_with_codistillation_param_change}, we compare the change in the norm of the parameters (since initialization) for models trained with codistillation (with constant weight decay as described in~\cite{anil2018large}) with the models trained with \allreduce. We observe that when training with \codistillation, the parameters remain closer to their initial values, thus demonstrating the regularizing effect of codistillation.

\subsubsection{Evaluating Codistillation For Different Models}

\begin{figure}[t]
\centering     %
\subfigure[ResNet50]{\label{fig:compare_ar_with_codistillation_resnet50_validation_top_1_accuracy}\includegraphics[width=0.48\textwidth]{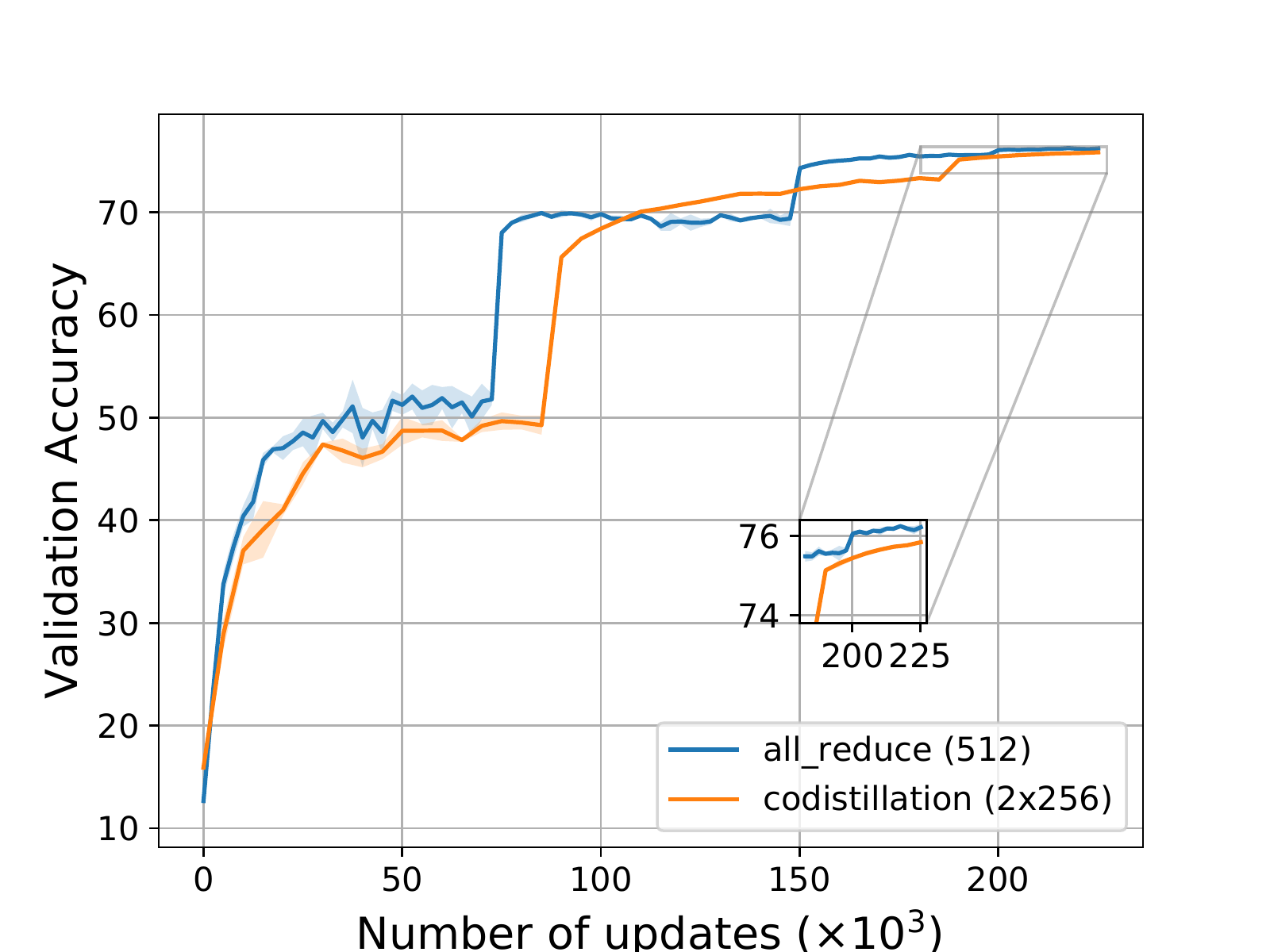}}
\subfigure[ResNeXt101]{\label{fig:compare_ar_with_codistillation_resnext101_validation_top_1_accuracy}\includegraphics[width=0.48\textwidth]{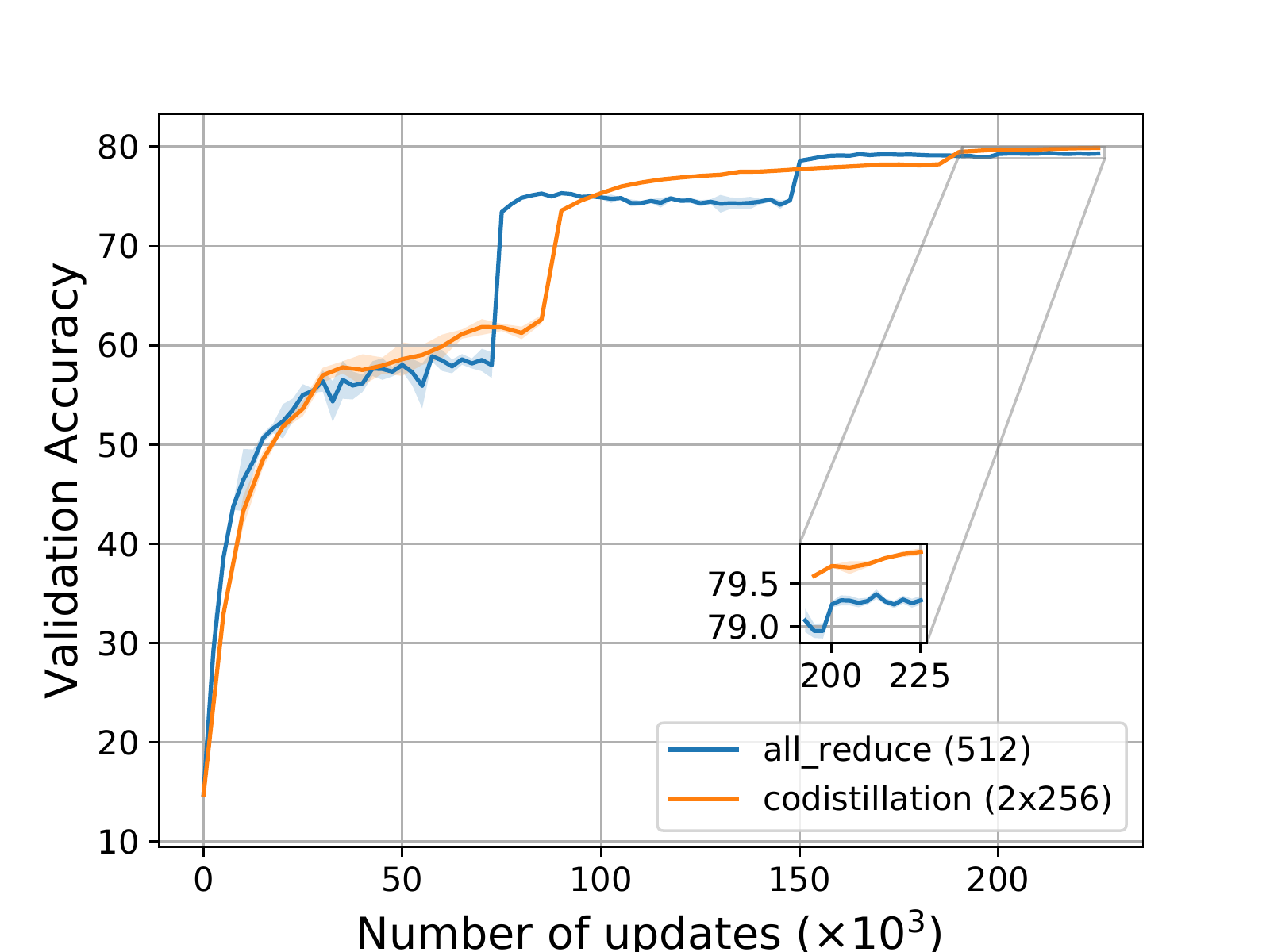}}
\caption{Comparing the top-1 validation accuracy for the \allreduce and \codistillation setups for ResNet50 and ResNeXt-101 models (respectively) on the ImageNet dataset, with decreasing weight decay and a shifted learning rate decay scheduled compared to~\cite{goyal2017imagenet_in_an_hour}. Both methods achieve similar values of top-1 validation accuracy. The corresponding training losses are shown in  Fig.~\ref{fig:compare_ar_with_codistillation_train_average_loss}.}
\label{fig:compare_ar_with_codistillation_validation_top_1_accuracy}
\end{figure}

\begin{figure}[h]
\centering
\subfigure[ResNet50]{\label{fig:compare_ar_with_codistillation_resnet50_train_average_loss}\includegraphics[width=0.48\textwidth]{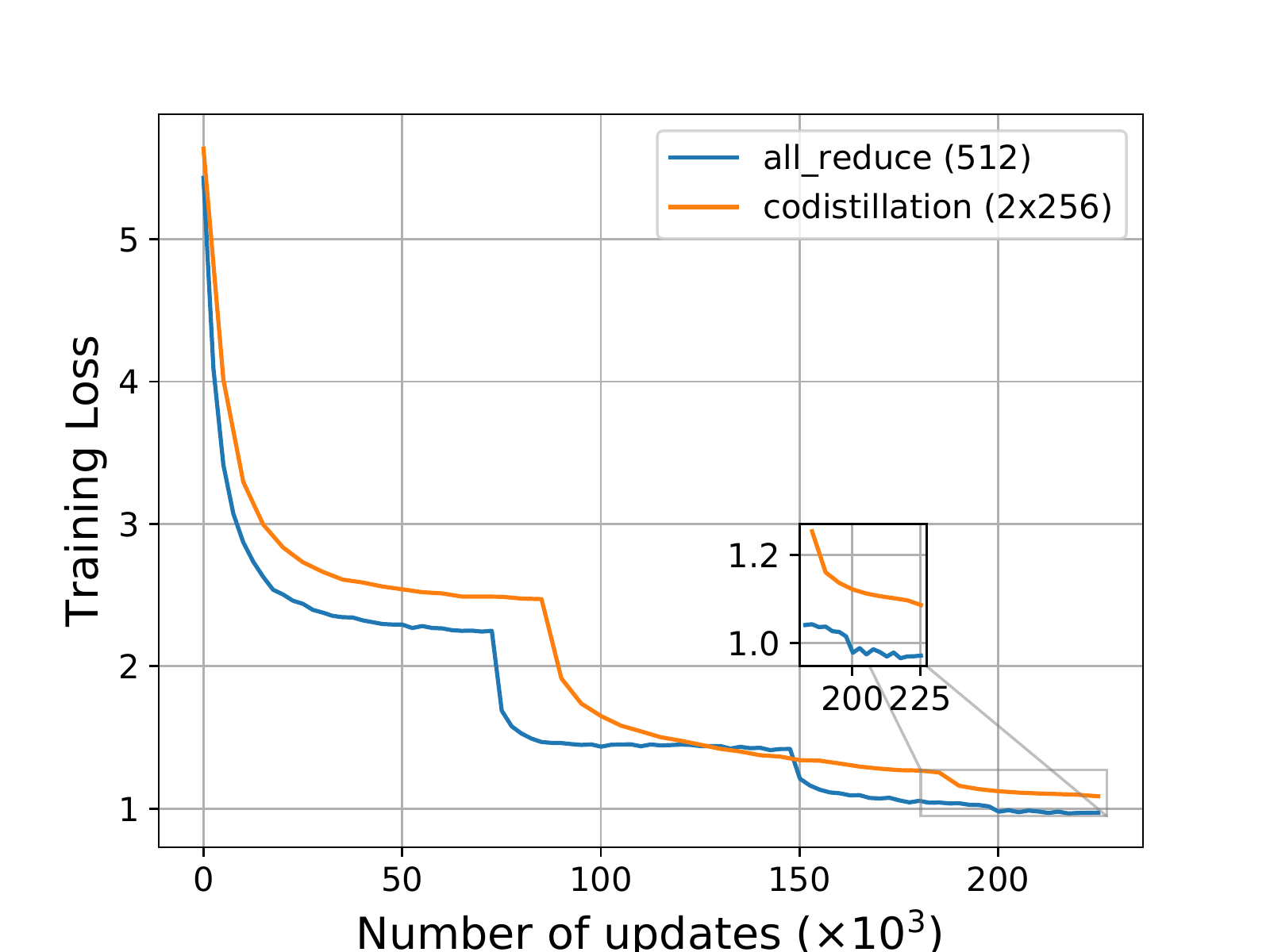}}
\subfigure[ResNeXt101]{\label{fig:compare_ar_with_codistillation_resnext101_train_average_loss}\includegraphics[width=0.48\textwidth]{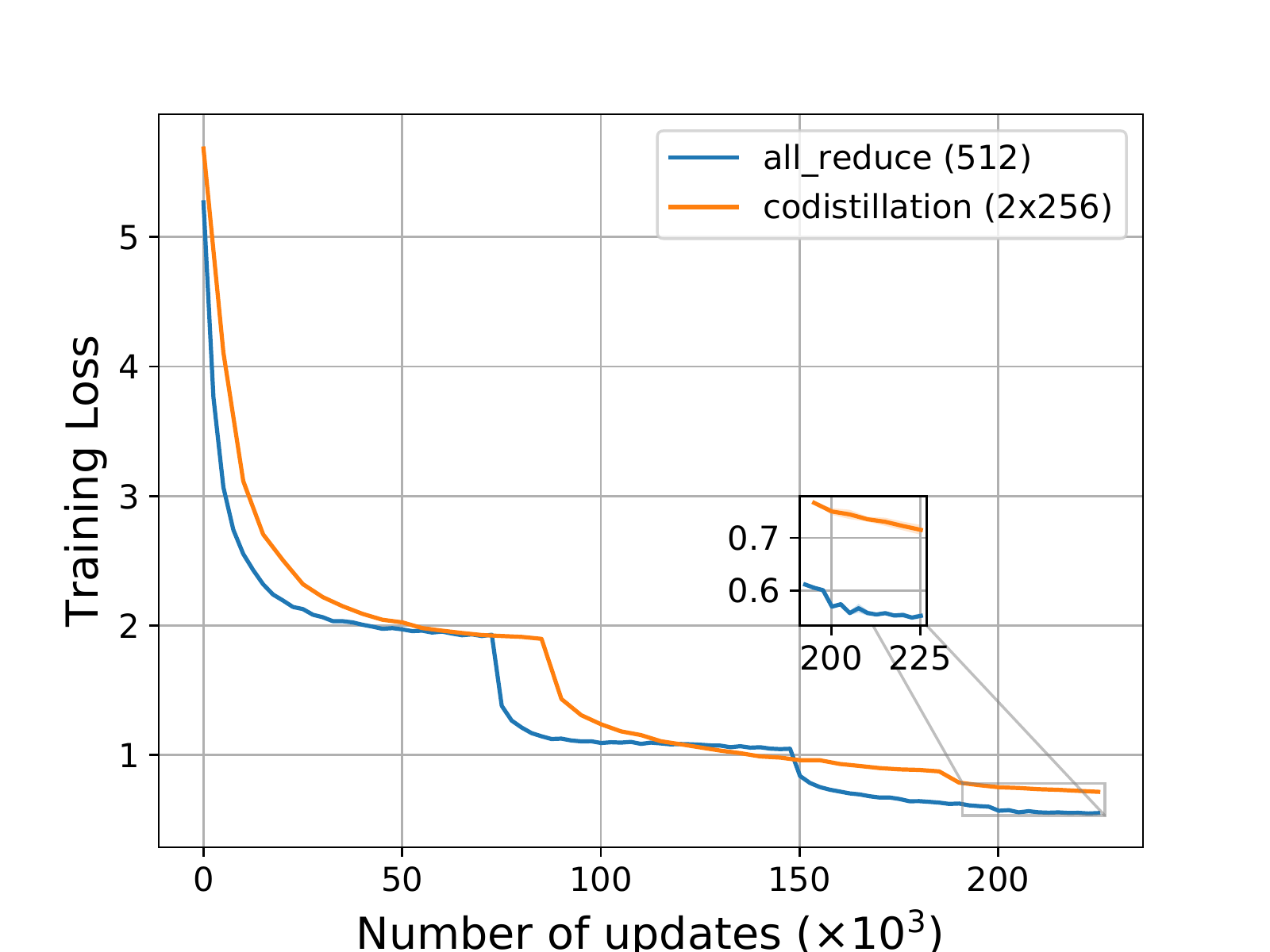}}
\caption{Comparing the training loss for the \allreduce and \codistillation setups for ResNet50 and ResNeXt-101 models (respectively) on the ImageNet dataset, with decreasing weight decay and a shifted learning rate decay scheduled compared to~\cite{goyal2017imagenet_in_an_hour}. While the \allreduce setup reaches a lower training loss, the performance on the validation dataset (in terms of top-1 validation accuracy) is very similar for the two setups (for both models) as shown in Fig.~\ref{fig:compare_ar_with_codistillation_validation_top_1_accuracy}.}
\label{fig:compare_ar_with_codistillation_train_average_loss}
\end{figure}

In Fig.~\ref{fig:compare_ar_with_codistillation_train_average_loss}, we plot the training loss for the ResNet50 (Fig.~\ref{fig:compare_ar_with_codistillation_resnet50_train_average_loss}) and ResNeXt101 (Fig.~\ref{fig:compare_ar_with_codistillation_resnext101_train_average_loss}) models for the ImageNet dataset, following the setup described in Section~\ref{app_subsec:codistillation_vs_large_batch_training} (based on the setup proposed in \cite{goyal2017imagenet_in_an_hour}) and accounting for the regularization effect. Specifically, \cite{goyal2017imagenet_in_an_hour} recommends using a constant $L2$ weight decay set to $10^{-4}$ throughout training. Keeping the initial value of this weight decay to $10^{-4}$, we reduce it to $10^{-5}$ after the first learning rate decay and further to $0$ after the second learning rate decay. We also modify the learning rate schedule from \cite{goyal2017imagenet_in_an_hour}, which is based on how the training loss changes (and saturates) during training. Due to the regularization effect of \codistillation, we observe that the model's training loss saturates slower and we shift the schedule by a few epochs to account for this (from 15, 30, 40 to 18, 38, 44). We note that for both models, the \allreduce setup reaches a lower training loss. Despite this, the performance on the validation dataset (in terms of top-1 validation accuracy) is very similar for the two setups (for both models) as shown in Fig.~\ref{fig:compare_ar_with_codistillation_validation_top_1_accuracy}.

\subsubsection{Evaluating Codistillation for Different Learning Rate Schedules}

\begin{figure}[h]
\centering     %
\subfigure[ResNet50]{\label{fig:compare_ar_with_codistillation_cosine_resnet50_train_loss}\includegraphics[width=0.48\textwidth]{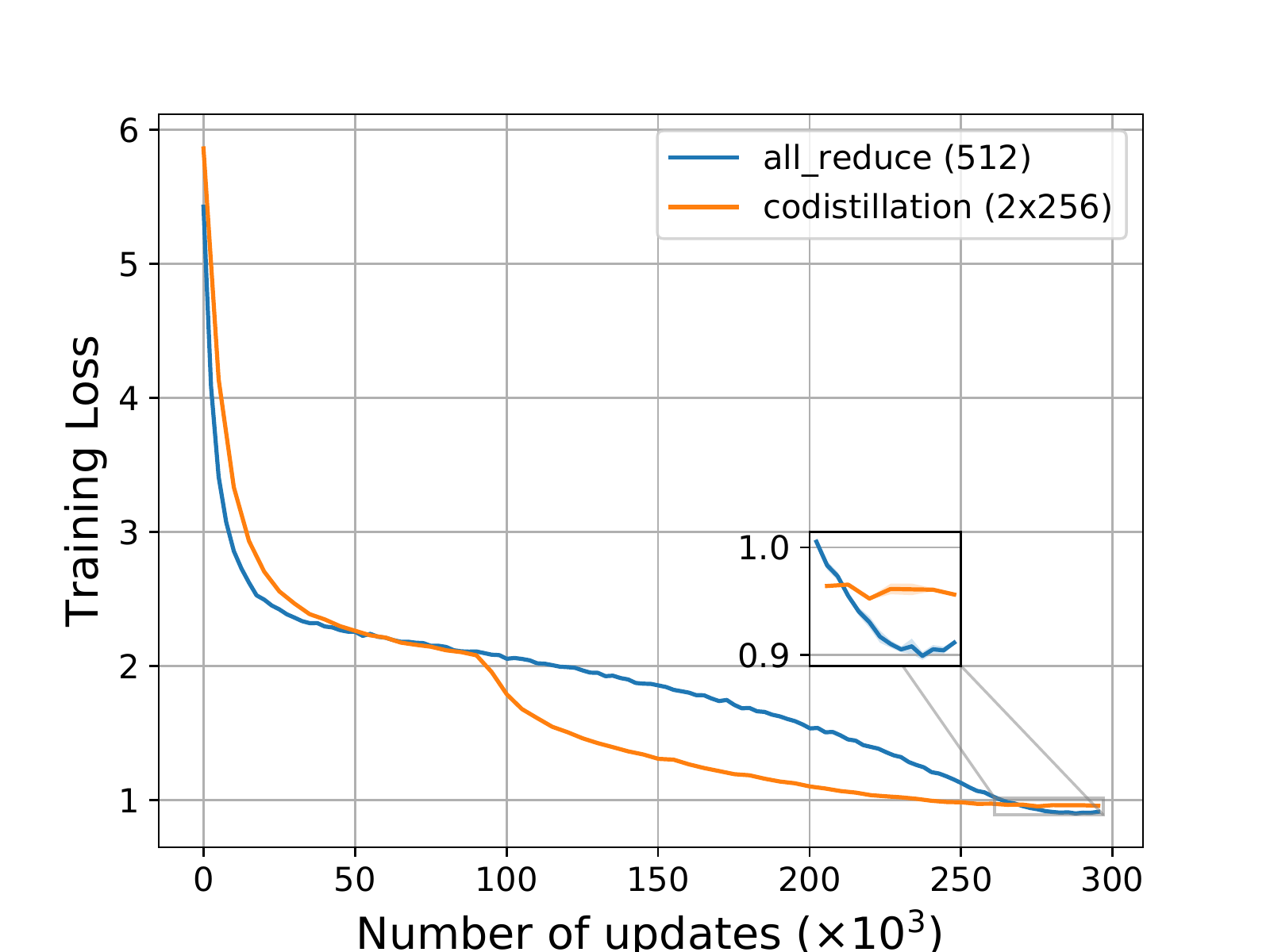}}
\subfigure[ResNeXt101]{\label{fig:compare_ar_with_codistillation_cosine_resnext101_train_loss}\includegraphics[width=0.48\textwidth]{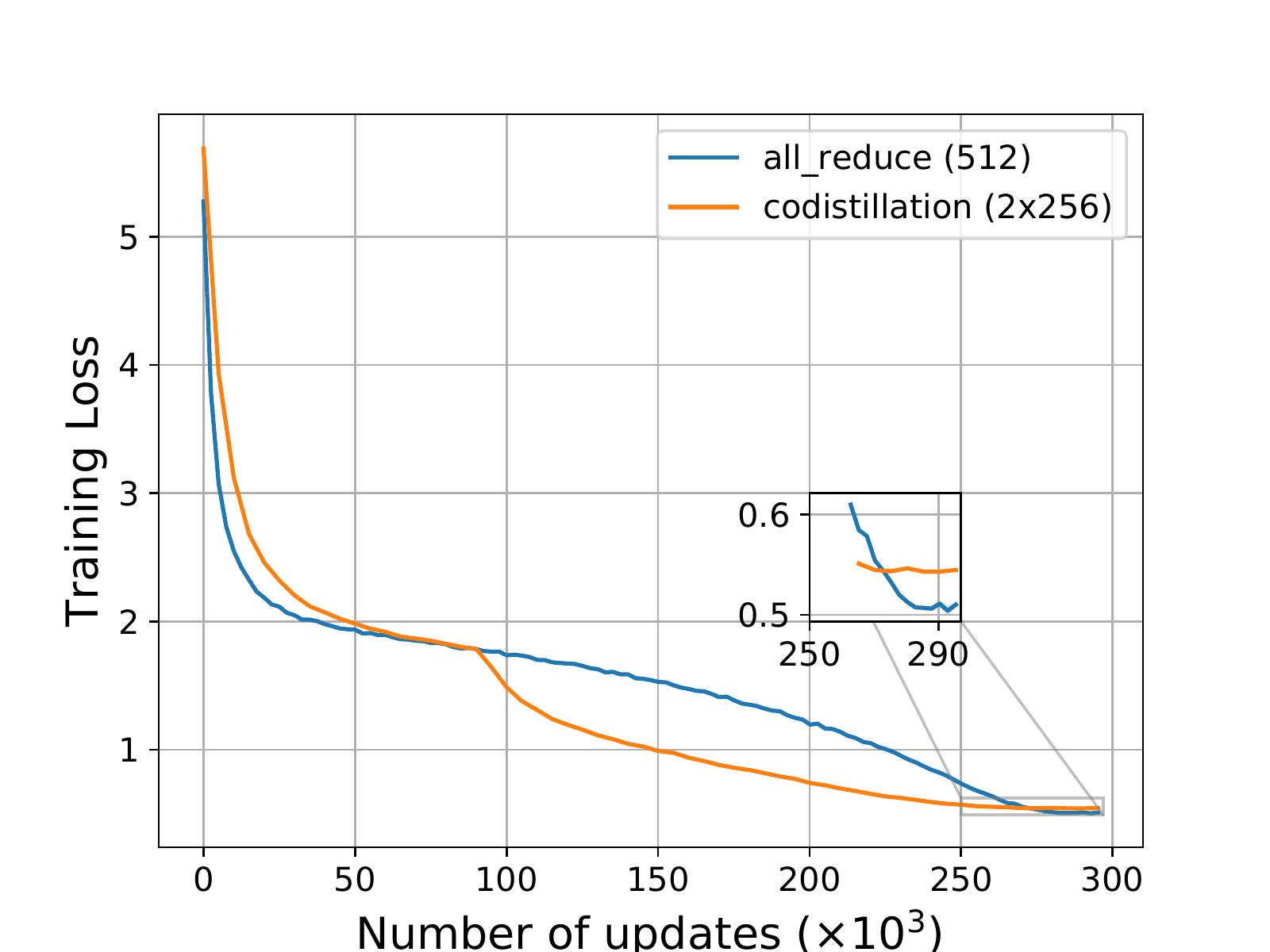}}
\caption{Comparing the training loss for the \allreduce and \codistillation setups for ResNet50 and ResNeXt-101 models (respectively) on the ImageNet dataset, using the cosine learning rate schedule proposed in \cite{he2019bag_of_tricks_for_image_classification_with_convolutional_neural_networks}. While the \allreduce setup reaches a lower training loss, the performance on the validation dataset (in terms of top-1 validation accuracy) is very similar for the two setups (for both models) as shown in Fig.~\ref{fig:compare_ar_with_codistillation_cosine_validation_top_1_accuracy}.}
\label{fig:compare_ar_with_codistillation_cosine_train_loss}
\end{figure}

So far, the ImageNet experiments in Section~\ref{app_subsec:codistillation_vs_large_batch_training} used the step-wise learning rate schedule described in \cite{goyal2017imagenet_in_an_hour}. We want to ascertain that our findings are not dependent on this specific learning rate schedule. Hence we train the ResNet50 and ResNeXt101 models with the cosine learning rate schedule~\citep{he2019bag_of_tricks_for_image_classification_with_convolutional_neural_networks}. In Fig.~\ref{fig:compare_ar_with_codistillation_cosine_validation_top_1_accuracy}, we observe that the final validation performance for the two approaches is very close, confirming that \codistillation works consistently across different learning rate schedules. The corresponding training loss plots are shown in Fig.~\ref{fig:compare_ar_with_codistillation_cosine_train_loss}.

\begin{figure}
\centering
\includegraphics[width=0.5\textwidth]{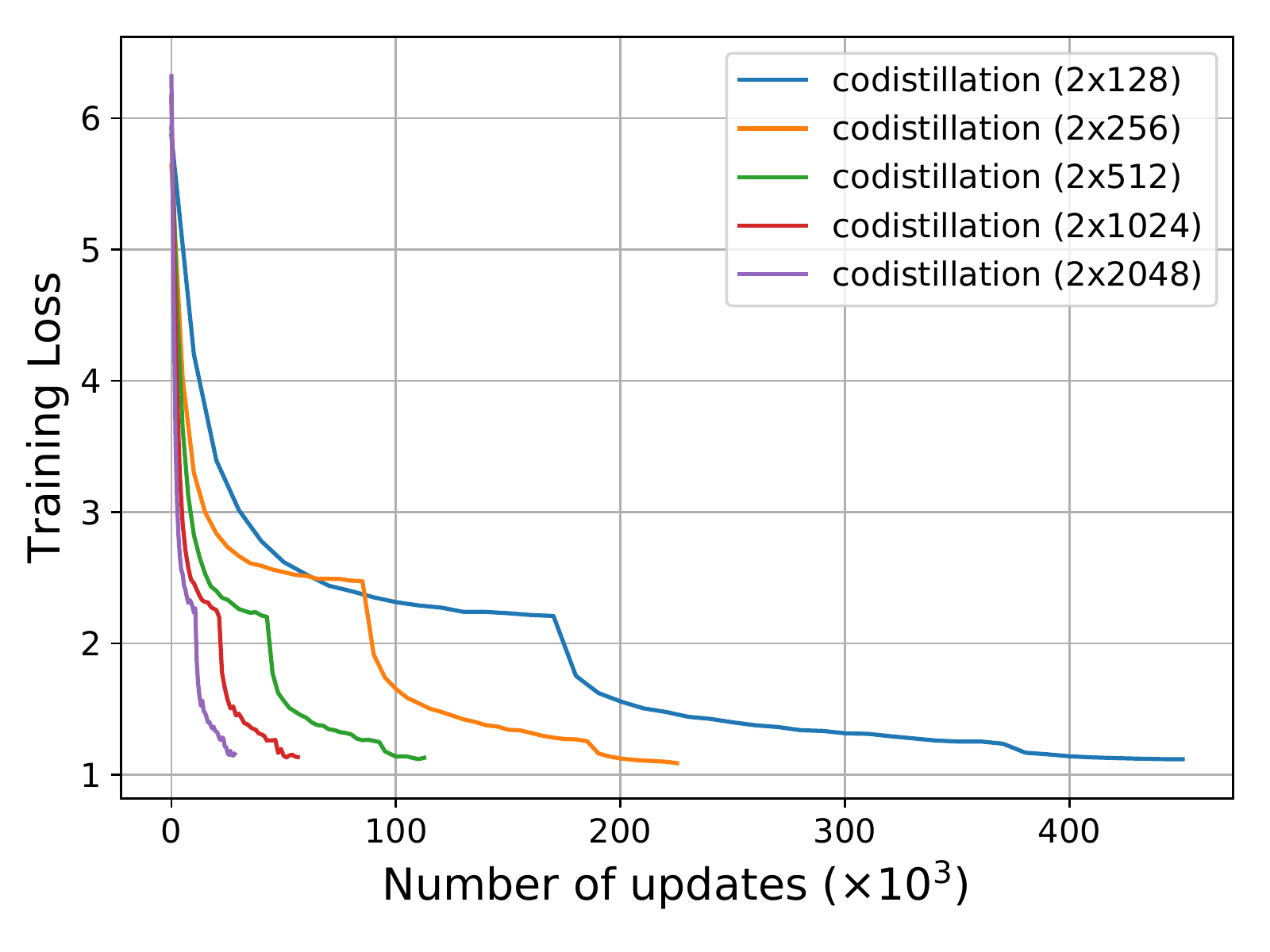}
\caption{Codistillation scales well across multiple values of batch size per worker. Each time we double the batch size per worker, we scale the learning rate schedule by a factor of two and perform half the number of updates. We do not observe any significant degradation in the training loss across a wide range of batch sizes. The training loss values are also reported in Table~\ref{tab:ar_using_different_number_of_gpus}.}
\label{fig:compare_codistillation_for_differernt_batch_sizes_resnet50_train_loss}
\end{figure}

\subsubsection{Scaling the Number of Workers for Codistillation}

In synchronous SGD (\allreduce), more workers can be added to increase the effective batch size (summed across all workers). The increased batch size reduces the gradient's variance and the model can be trained with a larger learning rate and fewer steps, while maintaining a similar level of accuracy~\citep{goyal2017imagenet_in_an_hour}. In Fig.~\ref{fig:compare_codistillation_for_differernt_batch_sizes_resnet50_train_loss}, Fig.~\ref{fig:compare_codistillation_for_differernt_batch_sizes_resnet50_validation_top_1_accuracy} and Table~\ref{tab:ar_using_different_number_of_gpus}, we demonstrate a similar effect with \codistillation. As one doubles the batch size per worker (and hence doubles the effective batch size), the learning rate can also be doubled and the model reaches similar performance in half the number of steps.

\begin{table}[]
\centering  
\begin{tabular}{|c|c|c|}
\hline
\textbf{Batch Size} & \textbf{Training Loss} & \textbf{Validation Accuracy} \\ \hline
$2 \times 128$      & 1.12                   & 75.61                        \\ \hline
$2 \times 256$      & 1.08                   & 75.82                        \\ \hline
$2 \times 512$      & 1.12                   & 75.51                        \\ \hline
$2 \times 1024$     & 1.13                   & 75.45                        \\ \hline
$2 \times 2048$     & 1.14                   & 75.26                        \\ \hline
\end{tabular}
\caption{Codistillation scales well across multiple values of batch size per worker. Each time we double the batch size per worker, we scale the learning rate schedule by a factor of two and perform half the number of updates. We do not observe any significant degradation in the training loss across a wide range of batch sizes. Fig.~\ref{fig:compare_codistillation_for_differernt_batch_sizes_resnet50_train_loss} and Fig.~\ref{fig:compare_codistillation_for_differernt_batch_sizes_resnet50_validation_top_1_accuracy} shows how the training loss and validation accuracy evolves over time respectively.}
\label{tab:ar_using_different_number_of_gpus}
\end{table}

\begin{table}[]
\centering  
\begin{tabular}{|c|c|c|}
\hline
\textbf{n} & \textbf{BLEU Score} \\ \hline
$1$     & 33.12   \\ \hline
$2$     & 33.20    \\ \hline
$4$     & 33.53    \\ \hline
$8$     & 33.94    \\ \hline
\end{tabular}
\caption{$n-$ way Codistillation improves performance on the IWSLT-14 German to English dataset as $n$ is increased while keeping the number of updates per device to be the same.}
\label{tab:iwslt_n_way_codistillation}
\end{table}

\subsection{Codistillation Works Beyond Vision}
\label{app:codistillation_works_beyond_vision}
So far, all our experiments have focused on the ImageNet dataset and the ResNet family of models. Next, we evaluate the \codistillation mechanism for neural machine translation (NMT). Specifically, we train the ``big'' transformer model~\citep{transformer_vaswani2017attention_is_all_you_need} (6 blocks in the encoder and decoder networks) on the WMT'16 En-De translation dataset, following the setup described in \cite{ott2018scaling_neural_machine_translation}. We explicitly reduce the amount of regularization (to account for regularization added by \codistillation) by removing the label smoothing loss. Reducing the explicit regularization is important for achieving performance comparable to \allreduce. This observation is in line with our previous observations on convolutional models.

\begin{figure}[t]
\centering     %
\subfigure[Validation Loss]{\label{fig:ar_vs_codistillation_transformer_wmt16}\includegraphics[width=0.4\textwidth]{figures/results/wmt/softmax/ar_vs_codistillation_val.pdf}}
\subfigure[Training Loss]{\label{fig:wmt_ar_vs_codistillation_train_nll}\includegraphics[width=0.4\textwidth]{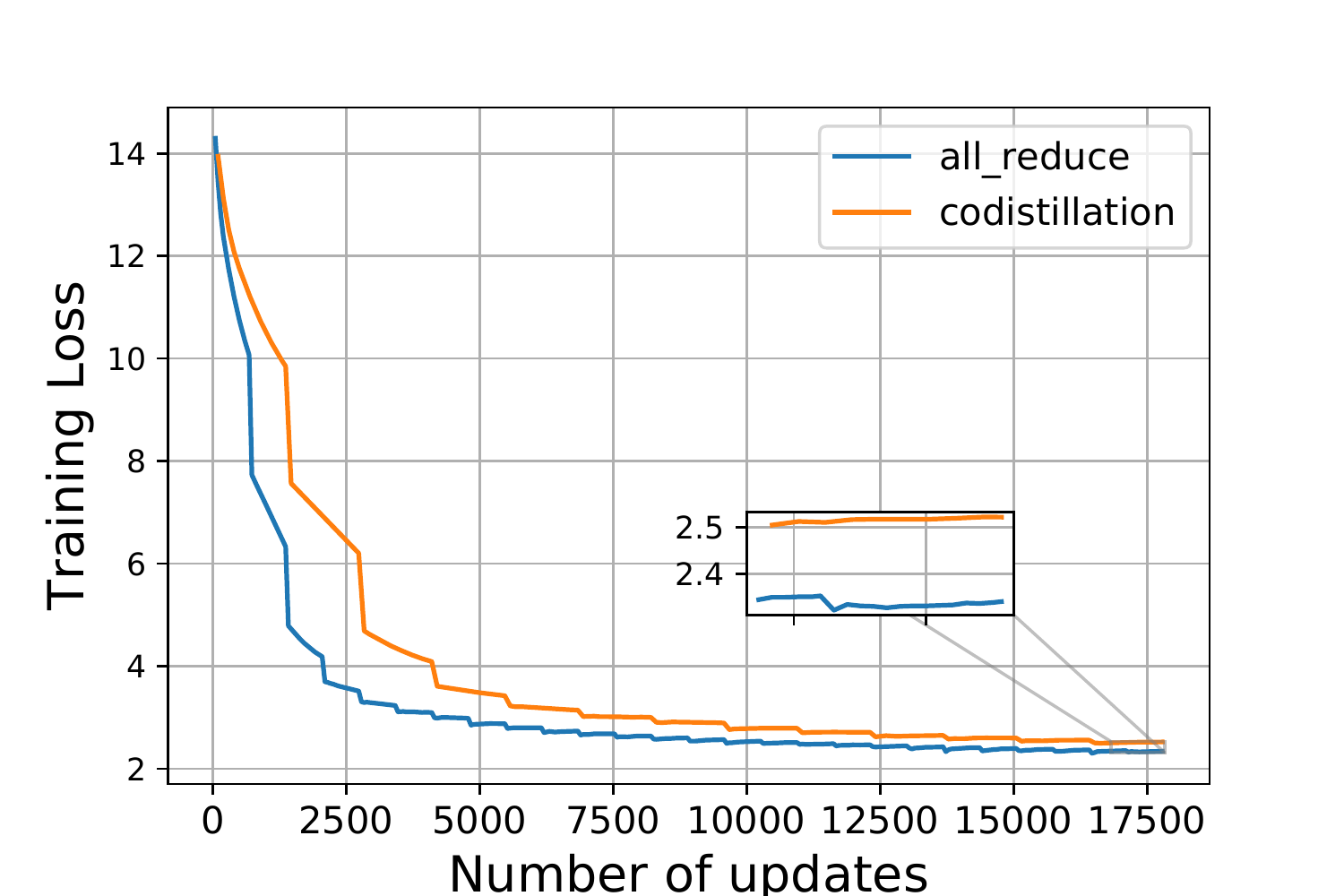}}
\caption{Comparing \allreduce and \codistillation using ``big'' transformer model on WMT'16 En-De dataset. The model trained using \codistillation performs worse in terms of training performance, but generalizes well to the validation dataset.}
\end{figure}

 In Fig.~\ref{fig:ar_vs_codistillation_transformer_wmt16}, we observe that the model trained with \codistillation reaches a similar validation loss as the model trained with \allreduce. This confirms that \codistillation with adjusted regularization schedule also extends to NMT. 

In Fig.~\ref{fig:wmt_ar_vs_codistillation_train_nll} we plot the training loss of the \allreduce and \codistillation setups on the WMT'16 En-De dataset.
We observe that \codistillation has a higher training loss, but almost matches the validation loss of \allreduce as shown in Fig.~\ref{fig:ar_vs_codistillation_transformer_wmt16}.

\begin{figure}[h]
\centering     %
\subfigure[Training Loss with ResNet50]{\label{fig:ar_vs_codistillation_variants_resnet50_train_loss}\includegraphics[width=0.48\textwidth]{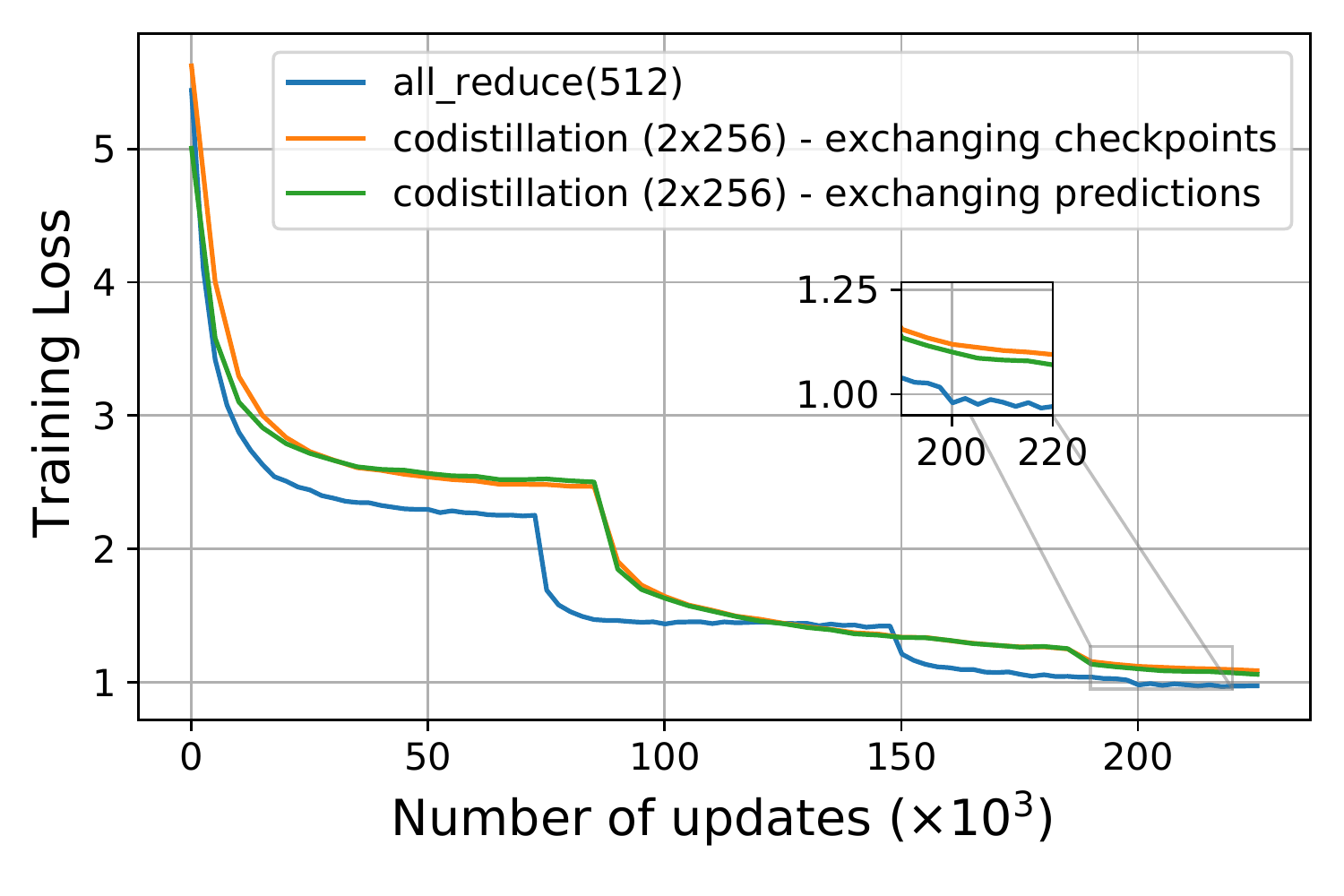}}
\subfigure[Validation Accuracy with ResNet50]{\label{fig:ar_vs_codistillation_variants_resnet50_validation_top_1_accuracy}\includegraphics[width=0.48\textwidth]{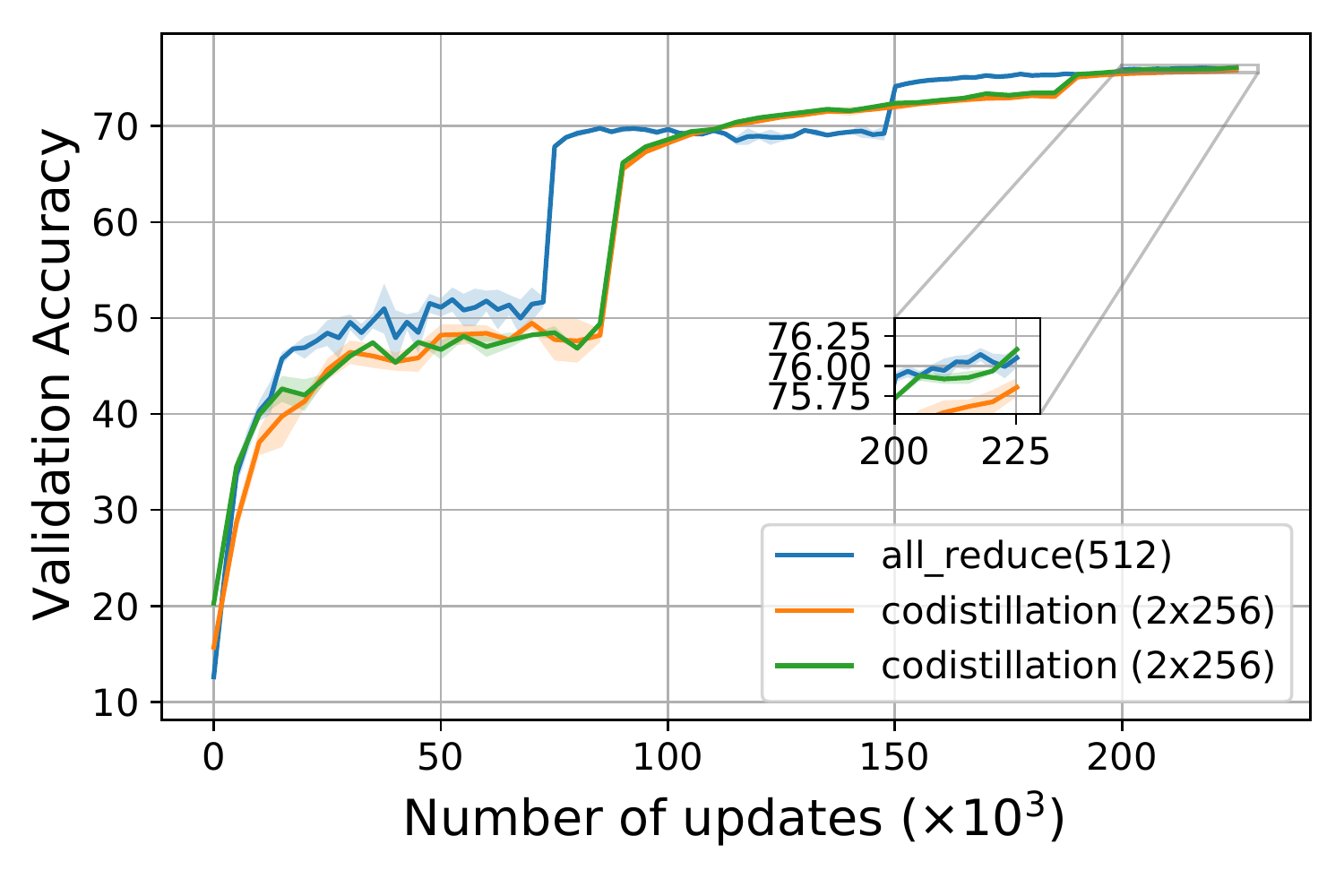}}
\caption{Comparing the training loss and validation accuracy (respectively) on the ImageNet dataset for training a ResNet 50 model using \allreduce and codistillation. We consider two variants of codistillation - (i) exchanging model replicas (or checkpoints), and (ii) exchanging model predictions. These two implementations are described in Section~\ref{sec:implemnentation_options}. We observe that codistillation can be competitive to \allreduce setup. In Fig.~\ref{fig:comparing_bits}, we show that codistillation requires communicating upto $\times 1000$ fewer bits. Taken together, the two observations highlight that codistillation requires communicating orders of magnitude  can be orders of magnitude fewer bits while obtaining comparable performance to \allreduce.}
\label{fig:ar_vs_codistillation_variants_resnet50}
\end{figure}

\subsection{Codistillation between multiple architectures}
\begin{figure}[h]
\centering     %
\subfigure[Training Loss with ResNet50]{\label{fig:codistillation_variants_train_loss}\includegraphics[width=0.48\textwidth]{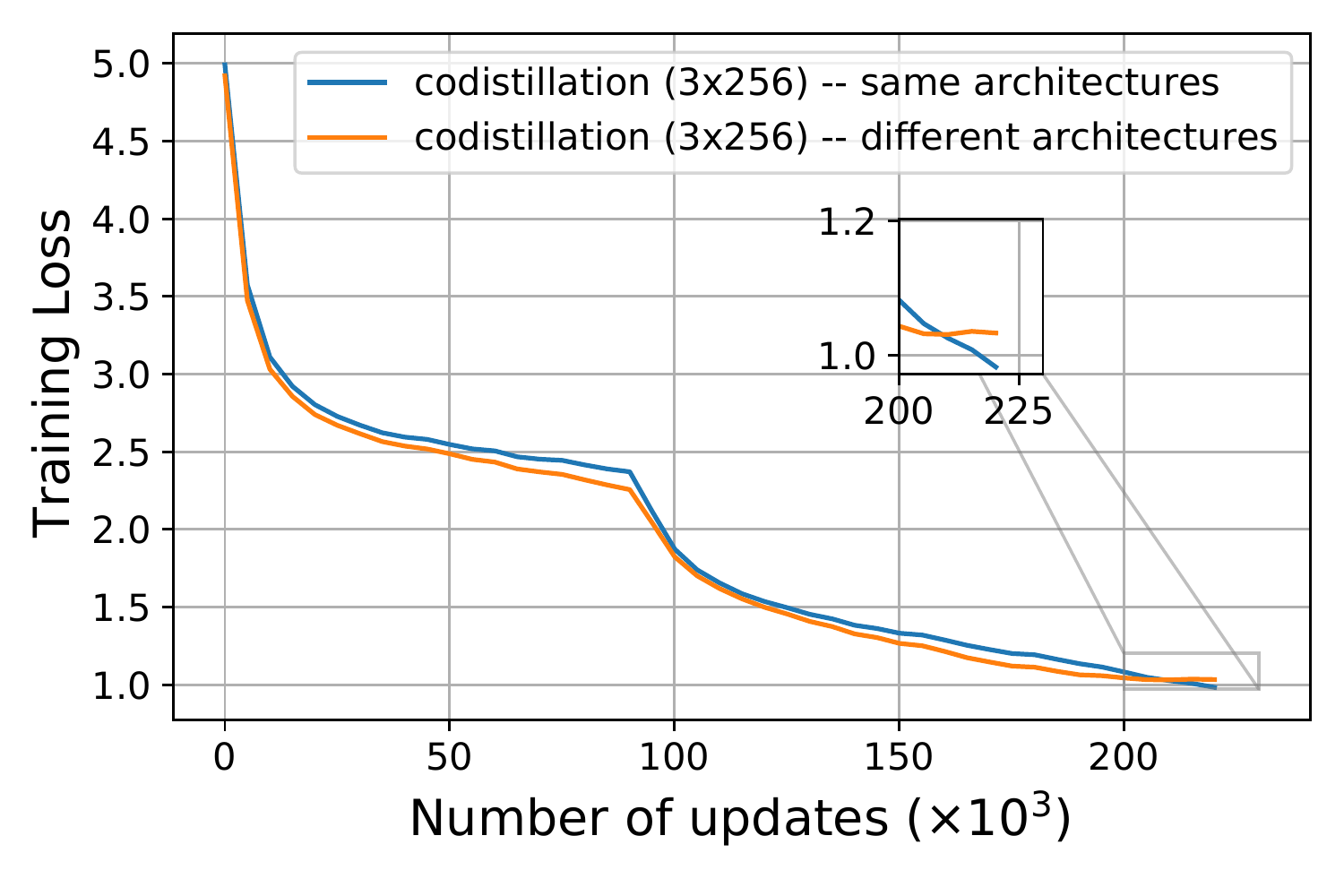}}
\subfigure[Validation Accuracy with ResNet50]{\label{fig:codistillation_variants_validation_accuracy}\includegraphics[width=0.48\textwidth]{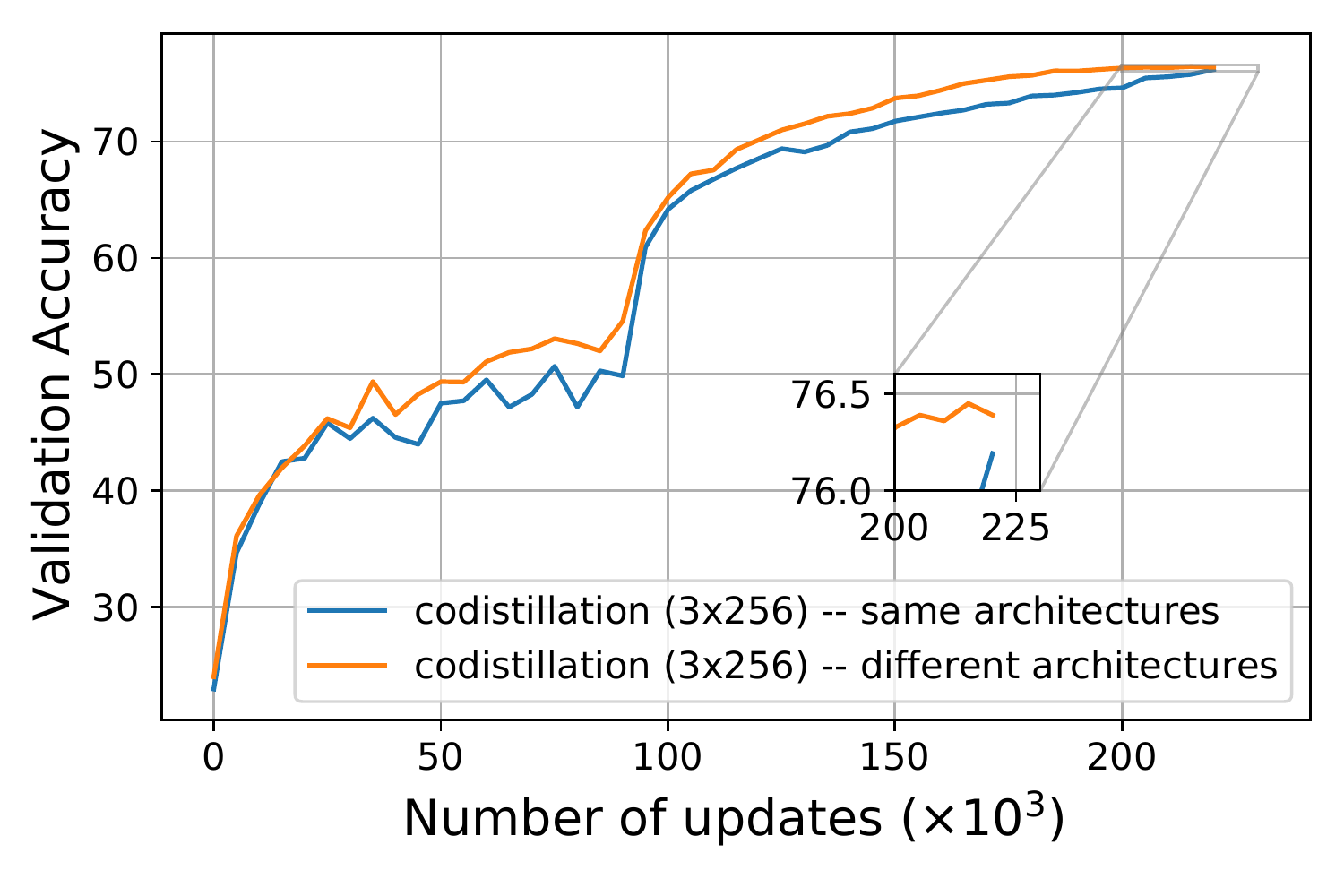}}
\caption{Comparing the training loss and validation accuracy (respectively) on the ImageNet dataset for $3-$way codistillation with same or different architectures.}
\label{fig:3_way_codistillation_between_different_architectures_3_way}
\end{figure}

\begin{figure}[h]
\centering     %
\subfigure[Training Loss with ResNet50]{\label{fig:codistillation_variants_train_loss_different_architectures}\includegraphics[width=0.48\textwidth]{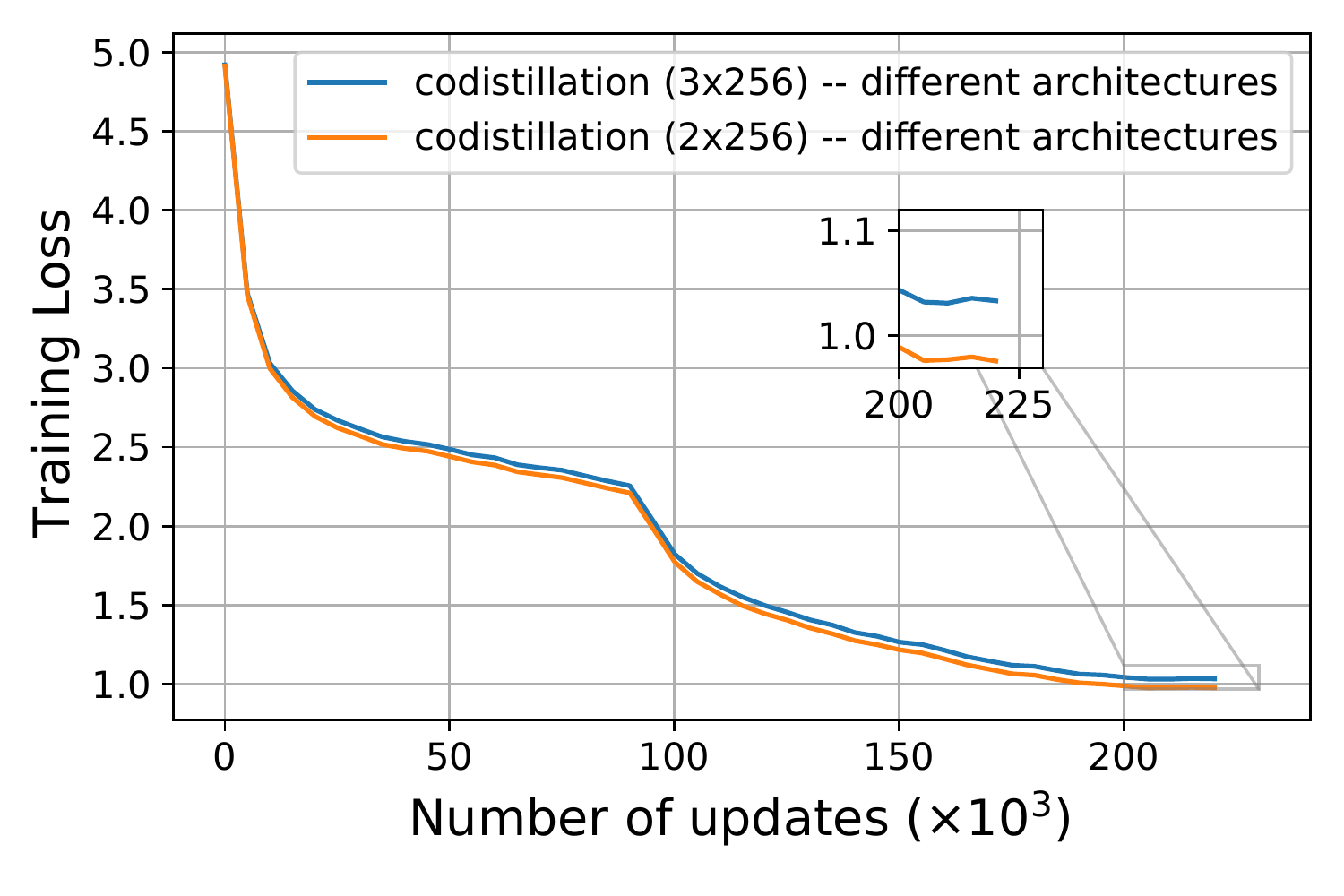}}
\subfigure[Validation Accuracy with ResNet50]{\label{fig:codistillation_variants_validation_accuracy_different_architectures}\includegraphics[width=0.48\textwidth]{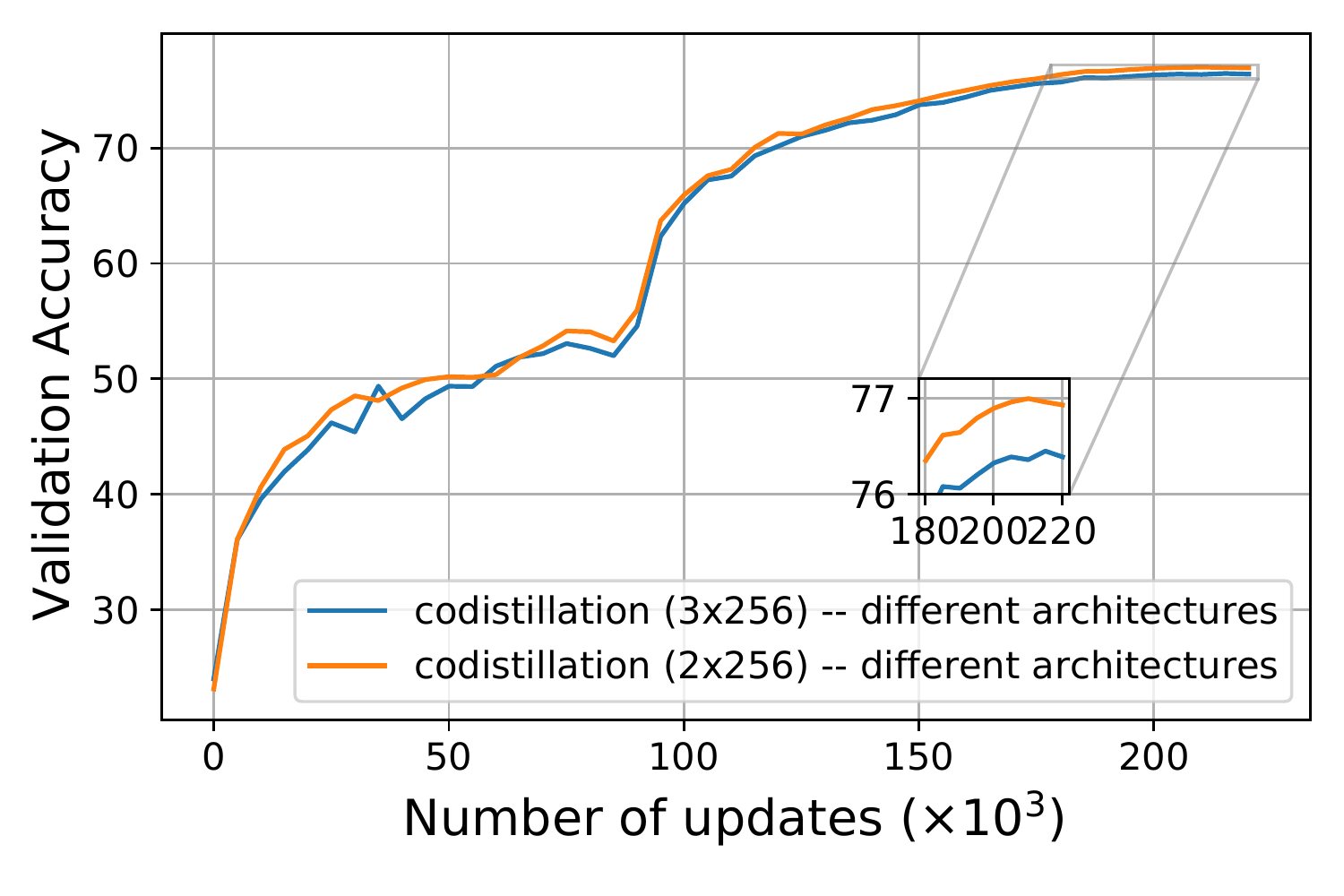}}
\caption{Comparing the training loss and validation accuracy (respectively) on the ImageNet dataset for $3-$way vs $2-$way codistillation with different architectures.}
\label{fig:3_way_codistillation_between_different_architectures_2_way_vs_3_way_different_architectures}
\end{figure}

\subsection{Codistillation Helps Reduce Overfitting}
\label{subsec:codistillation_helps_reduce_overfitting}

Section~\ref{app_subsec:codistillation_vs_large_batch_training} illustrated that applying codistillation in addition to other forms of regularization can result in over-regularizing, and progressively reducing the explicit regularization helps improve training and generalization. This section further explores the regularizing effect of codistillation by examining settings where we expect the model to overfit.

We simulate a scenario where the ResNet50 model is likely to overfit in a controlled way by training the model using only a fraction of the full ImageNet training set. When training on $(1/k)$th of the training set, we multiply the number of epochs by $k$ so that the total number of model updates performed is the same as when training with the full training set. The learning rate and weight decay schedules are also modified accordingly. As we train on less data, we expect the model to overfit, i.e., to obtain lower training loss and lower validation accuracy.

\begin{figure}
\centering     %
\subfigure[Final Validation Top-1 Accuracy]
{\label{fig:ar_vs_codistillation_sample_data_val_accuracy_agg}
\includegraphics[width=0.48\textwidth]{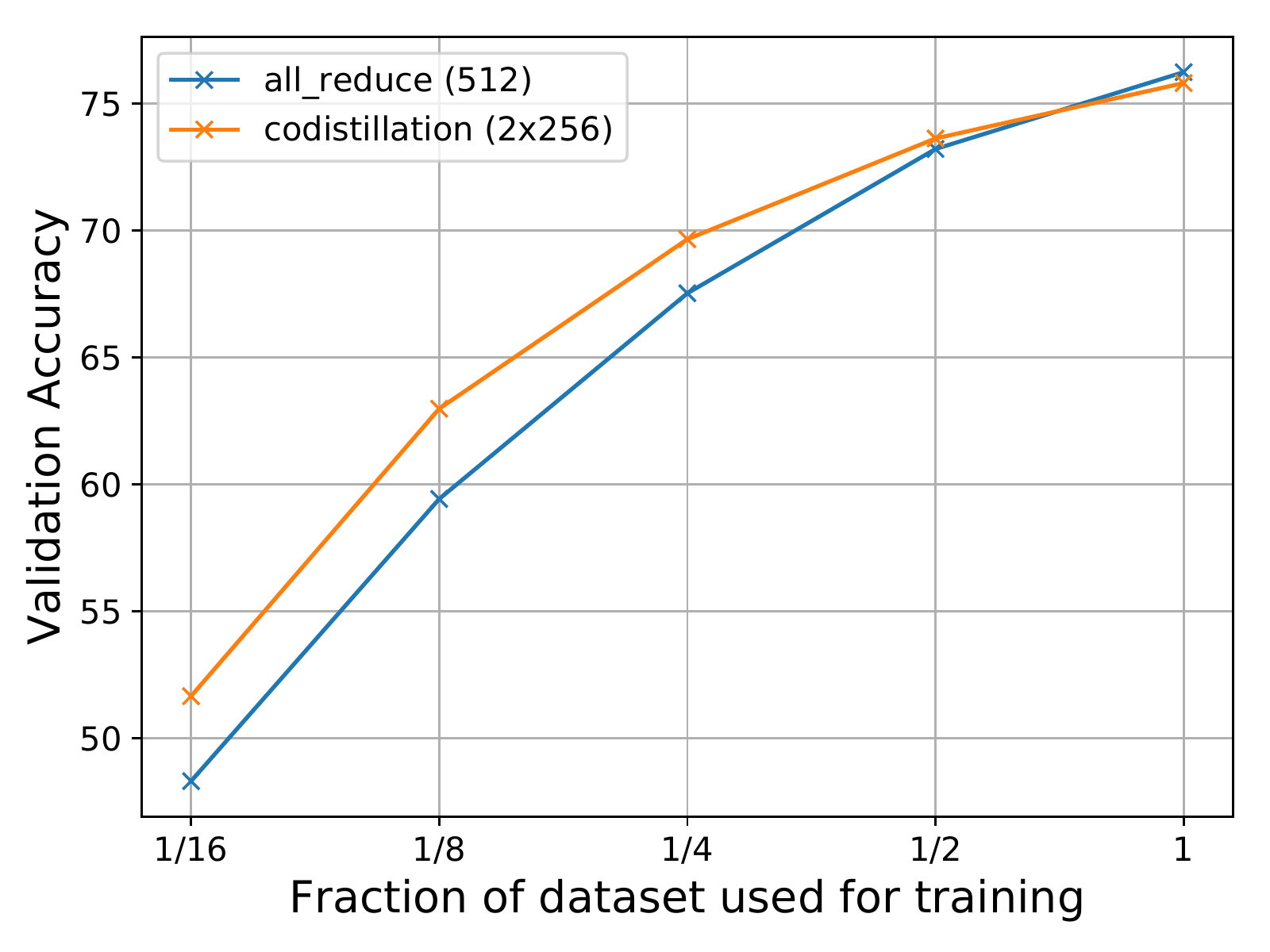}}
\subfigure[Final Training Loss]
{\label{fig:ar_vs_codistillation_sample_data_train_loss_agg}
\includegraphics[width=0.48\textwidth]{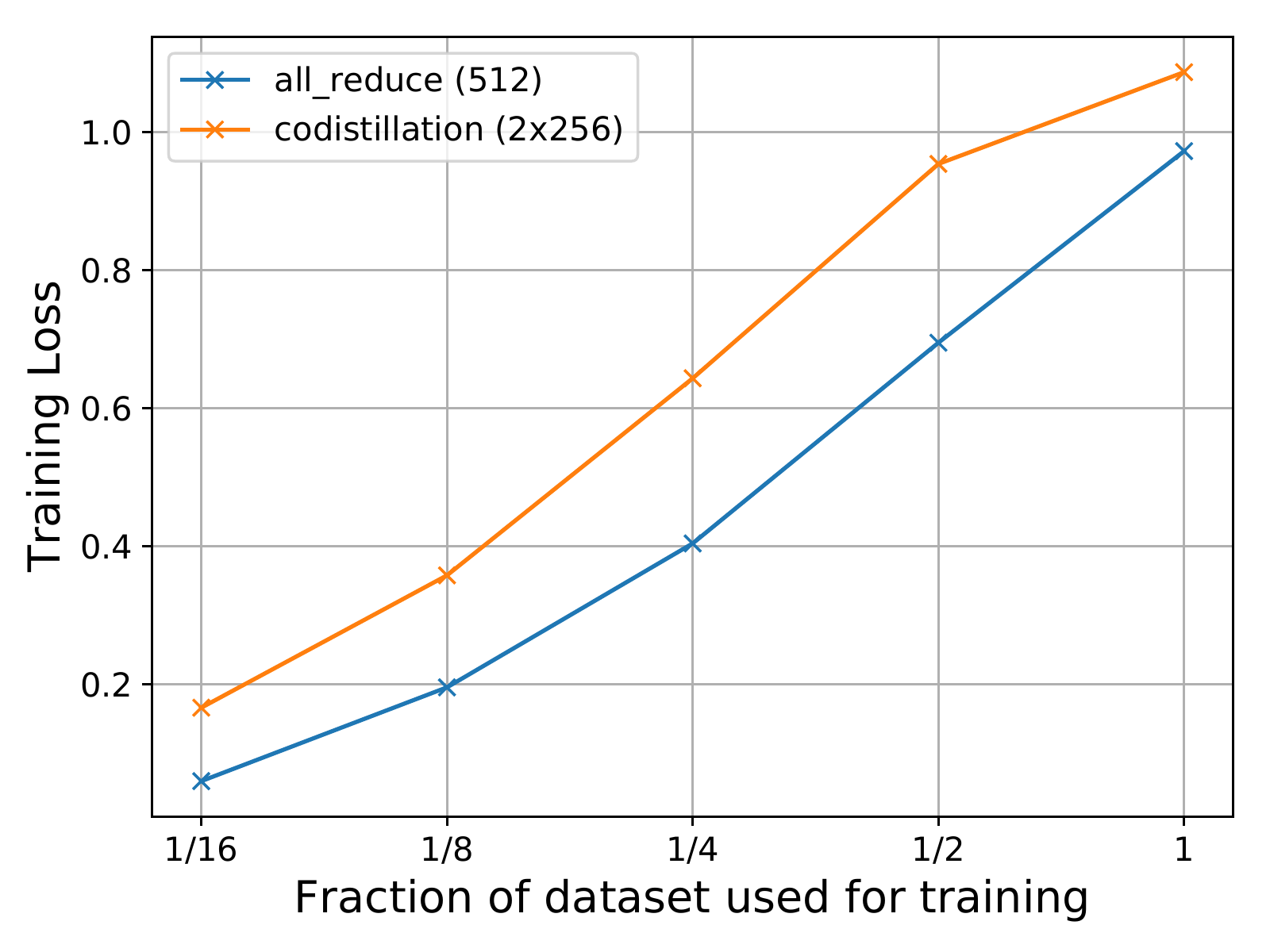}}
\caption{Final validation top-1 accuracy and training loss of a ResNet-50 model when trained using a fraction of the full training data. We observe as smaller fractions of training data are used (and model starts overfitting), \codistillation setup increasingly improves over the \allreduce setup in terms of validation accuracy.}
\label{fig:ar_vs_codistillation_sample_data_agg}
\end{figure}

In Fig.~\ref{fig:ar_vs_codistillation_sample_data_agg}, we indeed observe that overfitting occurs when using less training data. However, the overfitting is less severe when training using codistillation, providing further support for the hypothesis that codistillation acts as a regularizer.

An interesting side-effect of this observation is that codistillation could be an interesting alternative to \allreduce for training over-parameterized models. We verify this hypothesis by training a ``big'' Transformer model for a small NMT dataset (IWSLT 2014 German-English translation dataset that contains 153K training sentences, 7K development
sentences, and 7K test sentences). The model trained with codistillation achieves a validation NLL of 2.31 whereas the model trained with \allreduce reaches a validation NLL of 2.37.

\begin{figure}
\centering     %
\subfigure[Validation Accuracy]{\label{fig:scale_codistillation_to_3_models_with_fixed_compute_validation_accuracy}\includegraphics[width=0.48\textwidth]{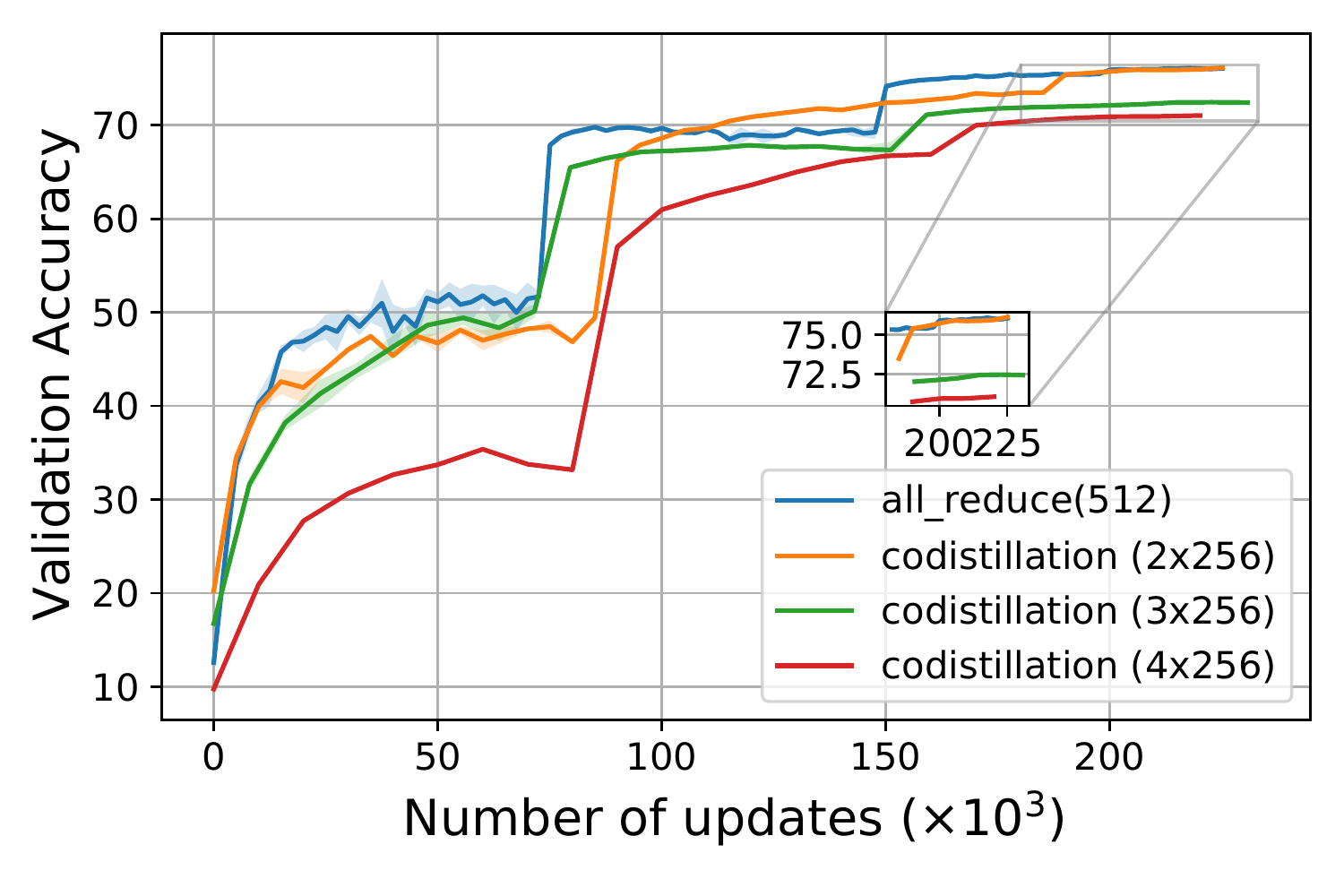}}
\subfigure[Training Loss]{\label{fig:scale_codistillation_to_3_models_with_fixed_compute_training_loss}\includegraphics[width=0.48\textwidth]{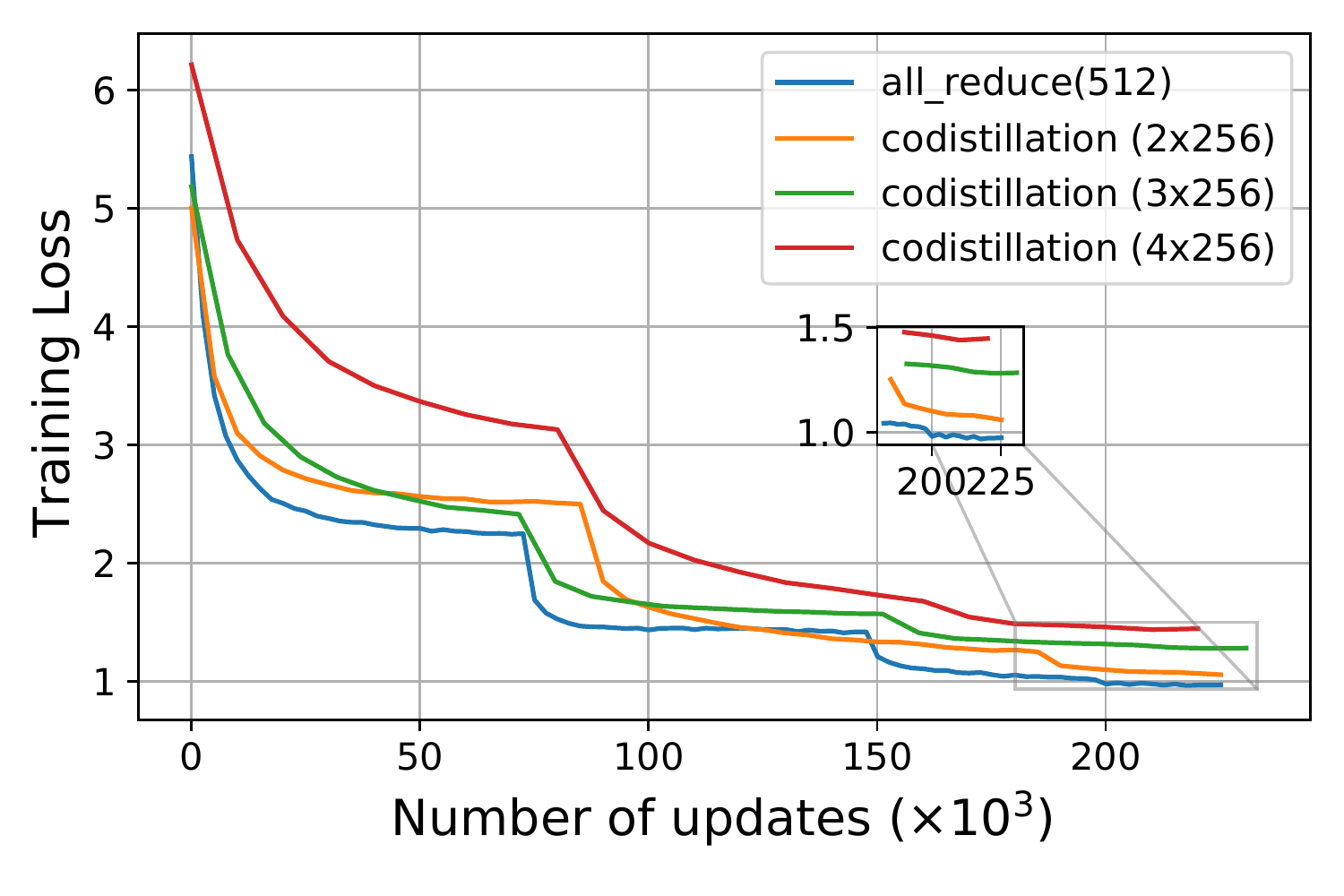}}
\caption{Effect of increasing $n$ while using a fixed number of update steps per model (i.e. reduced number of steps per device). As we increase $n$, the performance of codistilled model deteriorates.}
\label{fig:scale_codistillation_to_3_models_with_fixed_compute}
\end{figure}

\end{document}